\documentclass[runningheads]{llncs}

\usepackage{eccv}

\usepackage{eccvabbrv}

\usepackage{graphicx}
\usepackage{mwe}

\usepackage{booktabs}

\usepackage[accsupp]{axessibility}  

\usepackage[pagebackref,breaklinks,colorlinks,citecolor=eccvblue]{hyperref}

\usepackage{orcidlink}

\usepackage{dsfont}
\usepackage{booktabs}
\usepackage{array}
\usepackage{lipsum}
\usepackage{mathtools}
\usepackage{mwe}
\usepackage[inline]{enumitem}
\usepackage{tabularx}
\usepackage{siunitx}

\DeclareSIUnit{\B}{B}
\DeclareSIUnit{\count}{\relax}

\newcommand{\tinyscript}{\fontsize{6}{7}\selectfont}

\newcommand{\ours}{SVDtrunc\xspace}

\newcommand{\repeatthanks}{\textsuperscript{\thefootnote}}

\usepackage{url}

\usepackage{graphicx}
\usepackage{booktabs}
\usepackage{multirow}
\usepackage{float}
\usepackage{makecell}
\usepackage{subcaption}
\usepackage{amsmath}
\usepackage{blindtext}
\usepackage[dvipsnames]{xcolor}

\title{Importance-Aware Low-Rank Distillation of Diffusion Transformers}
\titlerunning{Importance-Aware Low-Rank Distillation}
\authorrunning{D.~Zavadski et al.}

\author{Denis Zavadski\thanks{Equal Contribution}\inst{1,2}\and
Sebastian Heid\repeatthanks\inst{1,2}\and
Damjan Kal\v{s}an\inst{1}\and
Stefan Roth\inst{2,3,4}\and
Carsten Rother\inst{1,2}}

\institute{
    \textsuperscript{1} Heidelberg University \quad \textsuperscript{2} Zuse School ELIZA \quad \textsuperscript{3}
    TU Darmstadt \quad \textsuperscript{4} hessian.AI
}

\begin{document}
\maketitle

\begin{abstract}
Diffusion Transformers (DiTs) have emerged as a dominant architecture for high-quality text-to-image generation, yet their scale poses challenges for efficient deployment. While truncated singular value decomposition (SVD) is a principled tool for parameter reduction, evidence from large language models (LLMs) suggests that naive low-rank approximation can cause catastrophic failure. In contrast, we find that truncated SVD in DiTs produces smooth degradation even under substantial global compression, with redundancy distributed across projection matrices throughout the whole network rather than concentrated in a few transformer blocks.
Building on these insights, we introduce \emph{\ours}, a two-step block-level compression scheme, first allocating ranks across blocks and compressing the least important ones via truncated SVD under a global parameter budget, and then fine-tuning all blocks with modular knowledge distillation and a rectified-flow objective.
We apply \ours to FLUX.dev across compression levels ranging from 40–\qty{90}{\percent} of the original parameter count. Across three benchmarks, GenEval, HPSv2, and DPG, we outperform all competing approaches. Notably, and in contrast to prior work, we retain near-full performance at \qty{68}{\percent} and remain competitive even at \qty{57}{\percent} of the original parameter budget. Furthermore, we show that \ours complements step distillation and achieves strong results even without fine-tuning, positioning it as a practical continuation of efficiency improvements beyond diffusion step reduction for large-scale generative models.
\newline\noindent Project page: \url{https://vislearn.github.io/SVDtrunc/}
\end{abstract}    
\section{Introduction}
\label{sec:intro}

    \begin{figure}[tb]
        \centering
        \includegraphics[width=.6\linewidth]{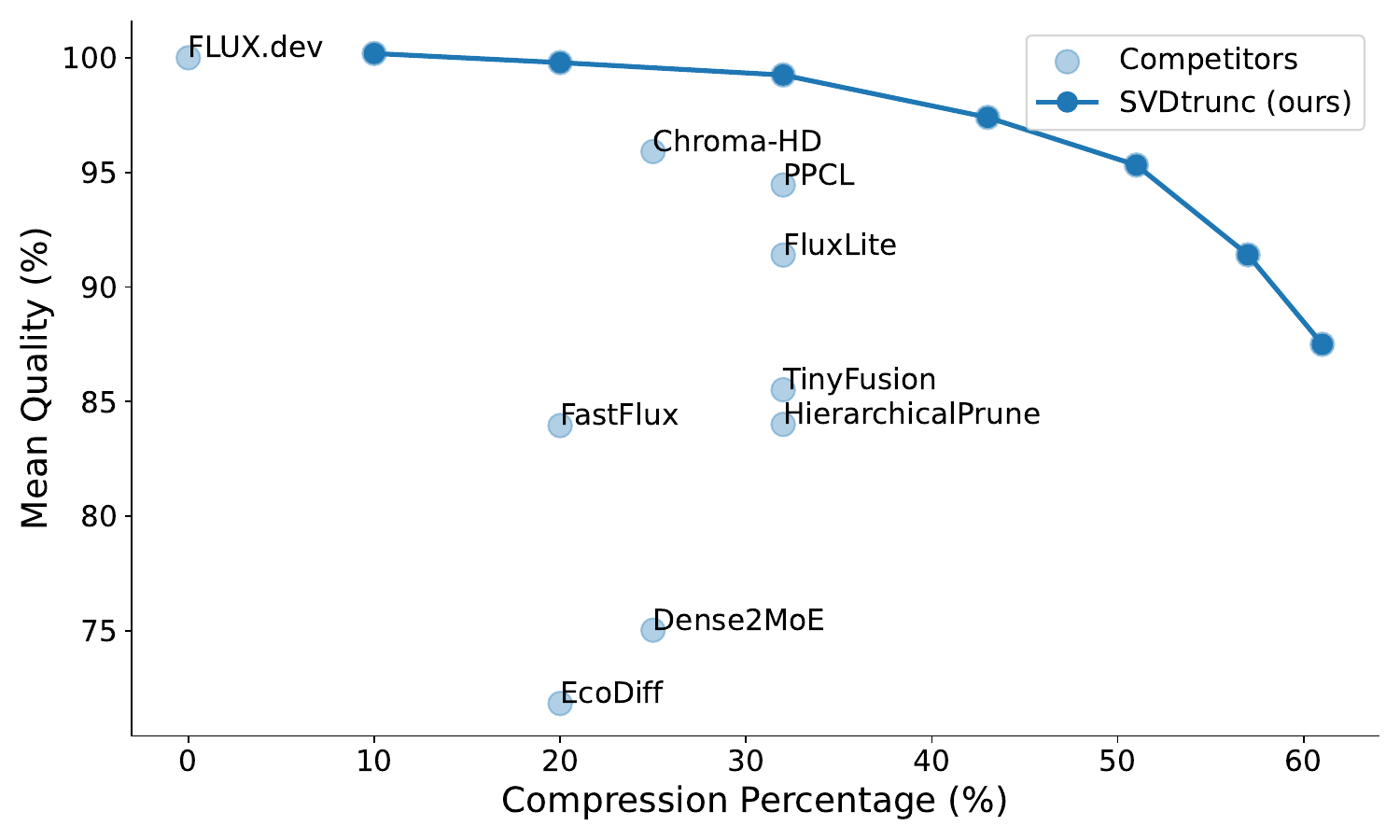}
        \caption{\textbf{Quantitative results} of our \ours approach compared to the state of the art. We plot quality relative to the original FLUX.dev model~\cite{flux2024} (in \%, $\uparrow$) against parameter reduction (\%, $\uparrow$). The quality is averaged across the relative performances on the GenEval\cite{GenEval}, DPG\cite{DPG}, and HPSv2\cite{HPSv2} benchmarks. We outperform prior models across all compression-levels.} 
        \label{fig:front_figure_plot}
    \end{figure}
    
Diffusion Transformers~\cite{DiT} (DiTs) have emerged as a dominant backbone for high-quality text-to-image generation. Their strong compositional reasoning and perceptual fidelity are enabled by large-scale Transformer architectures with substantial parameter and memory requirements. As state-of-the-art models continue to be scaled~\cite{sana, qwen_image, playground}, efficient deployment becomes increasingly challenging, motivating research on reducing neural function evaluations and model parameters. On the computational side, step distillation~\cite{salimans2022progressive, ADD, sdxl-lightning, sana_sprint} reduces the number of diffusion or flow evaluations, approaching the theoretical limit of single-step generation. Once this limit is reached, further efficiency gains within the same pre-trained architecture have to come from reducing the parameter footprint. Existing parameter reduction methods~\cite{tinyFusion, PPCL} typically rely on pruning or structural modification of Transformer blocks to achieve compactness.

A principled alternative to pruning and structural modification is low-rank approximation via singular value decomposition (SVD)\cite{LM_matrix_decompos, expl_lin_struct_decompos, FWSVD}, which provides the optimal rank-constrained approximation of a matrix under standard norms (\ie, all unitarily invariant norms \cite{mirsky1960symmetric}). While attractive for parameter compression, studies in autoregressive LLMs show that naive low-rank approximation can substantially degrade quality unless combined with careful fine-tuning or architectural adaptation \cite{svdllm, svdllm_v2, dobisvd, yuan2023asvd, FWSVD}. Whether Diffusion Transformers exhibit similar sensitivity remains unclear, as diffusion models iteratively refine latent representations over many steps where structural perturbations may accumulate.

    \begin{figure}[tb]
        \centering
        \includegraphics[width=.9\linewidth]{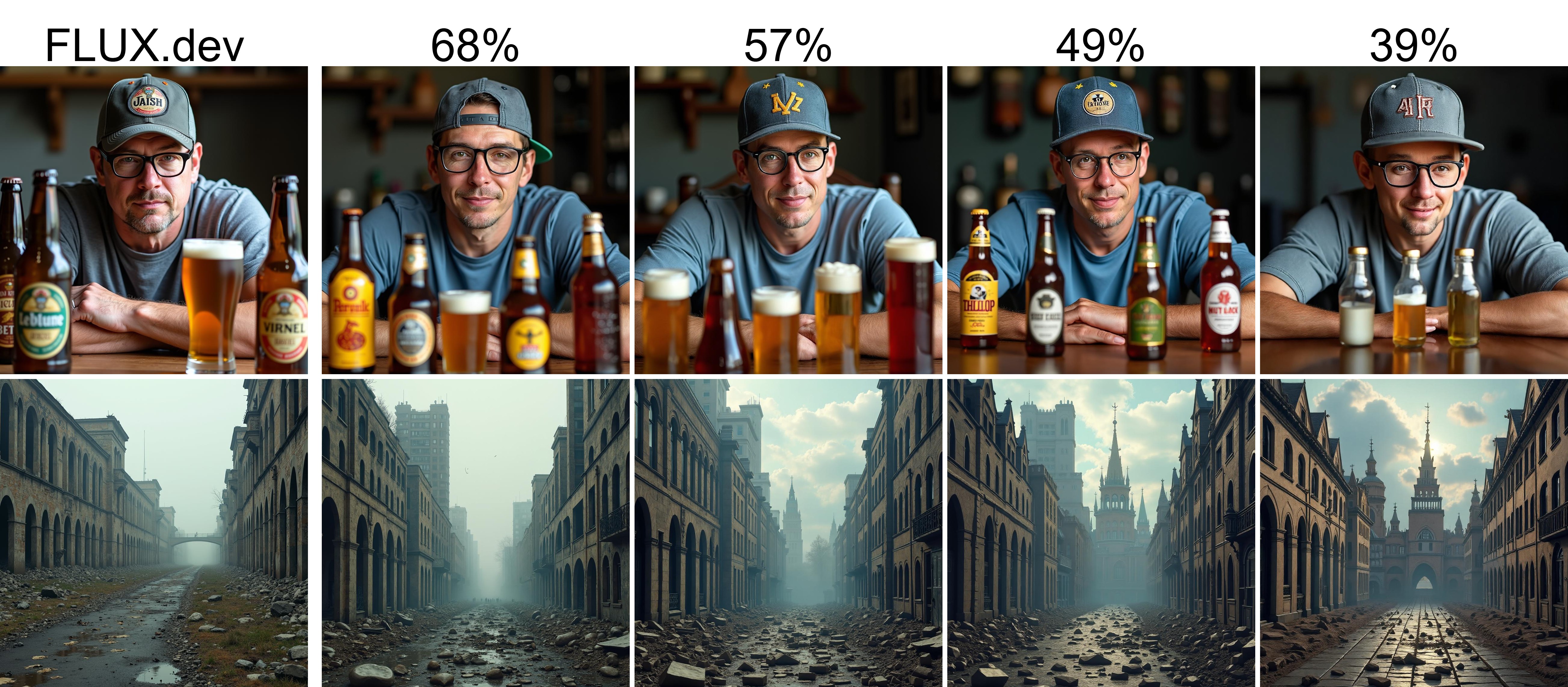}
        \caption{\textbf{Qualitative results} of applying our \ours compression schema to FLUX.dev\cite{flux2024}, ranging from \qty{68}{\percent} to \qty{39}{\percent} of the original parameter count. A minor degradation in quality is visible for \qty{39}{\percent} of parameters.}
        \label{fig:front_figure_images}
    \end{figure}

In this work, we show that Diffusion Transformers can be aggressively compressed via truncated SVD while retaining strong generative performance (\cref{fig:front_figure_images}). In contrast to LLMs, where naive low-rank approximation can cause substantial degradation and catastrophic failure \cite{svdllm, FWSVD}, we observe that DiTs remain surprisingly robust to rank reduction (\cref{fig:front_figure_plot}).
Even under large parameter reductions, compressed models preserve perceptual fidelity with only minor quality loss. This robustness emerges when compressing projection matrices rather than removing entire blocks, suggesting that redundancy in Diffusion Transformers can be exploited through distributed low-rank compression without architectural modification. To examine whether such redundancy is already present in pretrained DiTs rather than purely a result of distillation, we conduct a side experiment where models are compressed without fine-tuning. Even for substantial parameter reduction, \eg by up to \qty{30}{\percent}, low-rank approximation of projection matrices causes only minor quality degradation, whereas coarse architectural modifications lead to substantially stronger degradation \cite{cai2025fastflux, EcoDiff, tinyFusion}.

Prior work has shown that different Transformer blocks contribute unevenly to image quality, and that sensitivity varies across network depth~\cite{cai2025fastflux}. Building on this observation, we introduce a compression-based block importance estimation protocol, quantifying the functional contribution of individual blocks through controlled compression probing. We leverage this importance signal to allocate rank reduction non-uniformly given a global parameter budget. This enables to preserve critical components while compressing redundant ones more aggressively. Combined with modular knowledge distillation and rectified-flow training, this importance-aware rank allocation enables stable low-rank compression that preserves high representational capacity without altering the architecture.

Our contributions can be summarised as follows:
\begin{enumerate*}[label=\emph{(\arabic*)}]
    \item  We show that Diffusion Transformers exhibit strong intrinsic tolerance to projection-level low-rank approximation. In contrast to observations in autoregressive LLMs, truncated SVD induces gradual rather than catastrophic degradation, even in a training-free setting.
    \item  We provide empirical evidence that redundancy in DiTs can be exploited across projection matrices, as distributed low-rank compression preserves functionality significantly more gracefully than coarse block removal.
    \item  We introduce a block-level compression probing protocol for estimating block sensitivity and use it for importance-aware rank allocation, improving quality retention under matched budgets.
    \item  Our \ours establishes a new quality-parameter frontier for compressed DiTs and show that low-rank tolerance persists after step distillation.
\end{enumerate*}
\section{Related Work}
\label{sec:rel_work}

\textbf{Text-to-Image Generators.}
Text-to-image generation has evolved from U-Net-based\cite{Unet} latent diffusion models\cite{ldm} to large-scale Diffusion Transformers (DiTs)\cite{DiT, pixart_sigma, sana}. Early latent diffusion architectures such as SDXL~\cite{sdxl} reached approximately \qty{2.6}{\B} parameters, while recent DiT-based models, including FLUX\cite{flux2024} and Qwen-Image\cite{qwen_image}, scale from 8 to \qty{20}{\B} parameters. Although Transformer scaling improves generation quality and semantic fidelity, such model sizes impose substantial memory and runtime constraints, making tuning and deployment challenging without specialised engineering. Reducing computational and memory cost has, therefore, become a central research topic.

\textbf{Temporal Distillation.}
Diffusion and rectified-flow models generate images iteratively over tens of steps. A large body of work focuses on compressing this trajectory by reducing the number of model evaluations\cite{dm_to_gan, salimans2022progressive, sdxl-lightning, ADD}. Step distillation and consistency-based methods commonly compress multi-step generation into 4–8 steps, substantially lowering latency.
However, such temporal distillation leaves the backbone architecture unchanged. The parameter count and VRAM footprint remain identical, and inference cost is fundamentally bounded by single-step evaluation. Consequently, temporal distillation does not address memory constraints and is complementary to structural compression.

\textbf{Structural Compression.}
Several works directly modify diffusion backbones to reduce parameters. EcoDiff\cite{EcoDiff} introduces differentiable neuron masking with an end-to-end objective spanning all denoising steps. It removes attention heads and FFN neurons via hard-concrete relaxation and achieves up to \qty{20}{\percent} sparsity on FLUX with lightweight mask optimisation. Pruned neurons are permanently removed from all forward passes, yielding structural sparsity and reduced memory. HierarchicalPrune\cite{HierarchPrune} performs block-level removal based on a position-aware sensitivity analysis, reducing model depth. PPCL\cite{PPCL} identifies redundant layer intervals and prunes or replaces them during inference, reducing memory and computation by eliminating contiguous representational stages. FastFLUX\cite{cai2025fastflux} applies branch-level substitution: instead of deleting blocks, it replaces the residual branch of selected ResBlocks with lightweight linear layers while preserving shortcut connections, combining architectural modification with localised re-training. Dense2MoE\cite{dense2Meo} converts dense DiTs into mixture-of-experts architectures, reducing the number of activated parameters per forward pass. While this lowers effective computation, the total parameter count remains largely unchanged, and sparse routing introduces additional gating networks and training complexity.

Across these approaches, compression is achieved by removing or replacing architectural components (neurons, heads, residual branches, or entire blocks), thereby altering network structure and reducing representational capacity at a coarse granularity.Under aggressive compression regimes, such structural interventions may lead to abrupt performance degradation due to irreversible capacity reduction. In contrast, we use low-rank approximation to compress selected projection matrices while preserving the original architectural structure and pre-trained capabilities for all blocks, allowing for granular control.

\textbf{Low-Rank Compression.}
Low-rank factorisation has been widely explored for CNNs and large language models (LLMs)\cite{grasp, svdllm, dobisvd, yuan2023asvd, FWSVD, svdllm_v2}. In LLMs, SVD-based compression and related techniques are known to be highly sensitive. Naive truncated SVD can even lead to catastrophic failure\cite{svdllm}, requiring carefully tuned truncation schedules\cite{grasp, FWSVD}.
In contrast, we find that for Diffusion Transformers truncated SVD-based compression is not sensitive. Applying SVD compression of projection matrices yields graceful degradation even at aggressive compression rates while preserving the original architecture.

\section{Importance-Aware Low-Rank Distillation}
\label{sec:method}
\begin{figure}[tb]
    \centering
    \includegraphics[width=.8\linewidth]{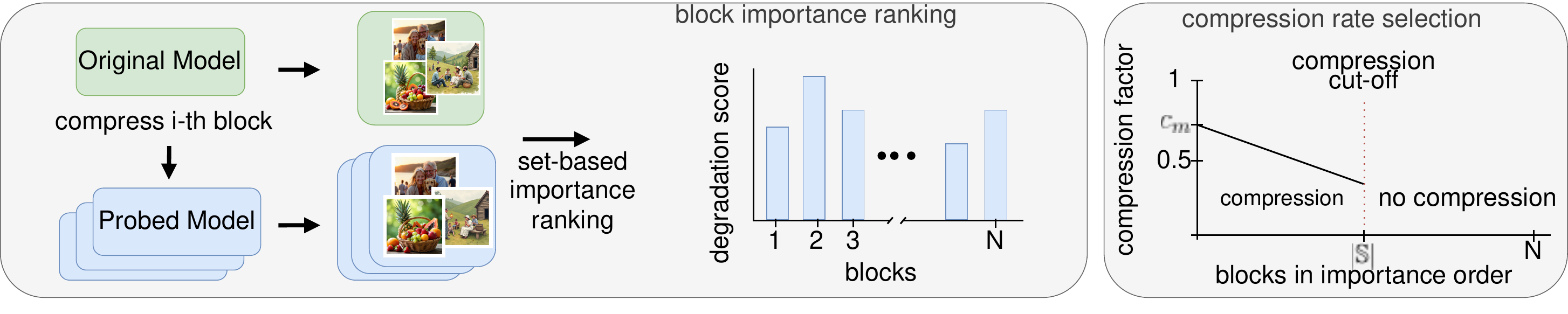}
    \caption{\textbf{\ours Overview.} \textit{(left)} We first generate $N$ probed models, by individually compressing each of the $N$ blocks. Then, each probed model generates a set of images.     
    By comparing these sets with the generations of the original model, we obtain a quality-based ranking of block importance. We use this ranking \emph{(right)} to assign importance-aware compression factors to each block of the final compressed model.}
    \label{fig:method}
\end{figure}
We base \ours on two hypotheses:  
\emph{(1)} Individual building blocks of the network are not fully redundant. Removing an entire block eliminates its functionality, which must then be compensated by the remaining blocks and may limit their performance.  
\emph{(2)} Building blocks do not contribute equally to the final generated image, implying that the degree of parameter reduction should vary across the network.  
Both hypotheses are empirically validated in \cref{sec:experiments}.

Typical parameter reduction methods prune entire Transformer blocks \cite{cai2025fastflux, dipgo}, a highly aggressive form of parameter reduction that violates our first hypothesis. Instead, to preserve block functionality, we compress individual blocks using truncated singular value decomposition (SVD) rather than removing them entirely. To address our second hypothesis, we introduce a block-importance-aware schedule that assigns different compression rates to blocks based on their estimated contribution to generation quality. Finally, we employ a modular knowledge distillation (MKD), similar to \cite{KDD}, to reproduce the features of all teacher modules. \Cref{fig:method} provides an overview of the truncated SVD procedure.

\subsection{Block Importance Analysis}
\label{subsec:method_block_analysis}

Previous work has studied how different parts of generative networks contribute to the generation and understanding of concepts and styles \cite{cnxs, HierarchPrune, deltadit}. In the context of compressing pre-trained text-to-image models, we focus on the importance of individual network blocks for the final output. To this end, we randomly sample $K$ images from an open-world dataset, caption them using a trained image captioner \cite{joycaption}, and generate one image per caption with the original, uncompressed model. We then probe the network by compressing one block at a time while keeping all others unchanged (compression details are given below). Images are generated again with identical prompts and seeds. For each of the $N$ probed blocks, we compare the quality of the resulting images to the original images using a quality metric $\mathcal{M}$.
\begin{figure}[tb]
    \centering
    \includegraphics[width=.6\linewidth]{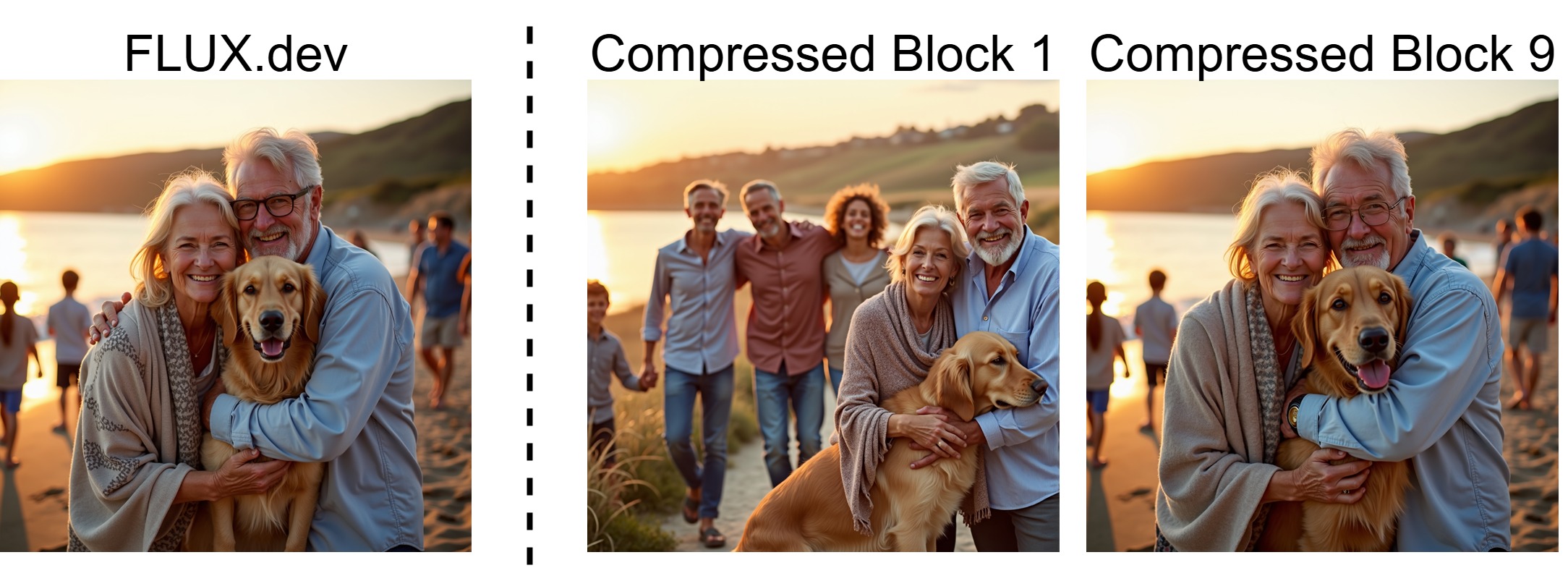}
    \caption{\textbf{Ranking metric justification.} Depending on which block is compressed, the generated image of the compressed model \emph{(right)} can deviate from the image generated by the uncompressed one \emph{(left)}. However, the pixel-wise change is not necessarily correlated with quality, making pixel-wise similarity scores less suitable for block ranking.}
    \label{fig:cmmd_validation}
\end{figure}

Since all images are generated under identical conditions, the generated sets contain corresponding images. This admits two choices for the metric $\mathcal{M}$: pairwise measures that compare each probing image with its original counterpart, \eg LPIPS \cite{LPIPS}, or set-level measures such as FID \cite{FID} or CMMD \cite{CMMD}. We empirically observe that compressing certain blocks can alter attributes such as pose or appearance, which affects pairwise scores like LPIPS. \Cref{fig:cmmd_validation} illustrates this effect for two compressed blocks of FLUX.dev~\cite{flux2024}. As such changes do not necessarily imply lower image quality, we argue that distribution-based metrics like CMMD or FID, which compare feature distributions across sets, are better suited for our importance analysis. Due to its greater stability with respect to the number of images, we adopt CMMD over FID for evaluation. This procedure yields a ranking of blocks according to their importance for generation quality, enabling stronger compression of blocks whose modification causes the least degradation.

\subsection{Importance-Aware Low-Rank Compression}
\label{subsec:method_importance_aware_compression}
Similar to previous approaches that aim to adapt or modify model architectures in a minimally destructive way to increase efficiency\cite{cai2025fastflux, cnxs, PPCL}, we reduce the parameter size of the model by reducing parameters on a whole model-block level. This means for modern DiT-based architectures~\cite{sd3_mmdit, flux2024, li2025omniflow, playground, qwen_image} that the considered blocks are single- and double-stream blocks, as shown in the supplement. These blocks are constructed from a variety of different projection layers and normalisations. When we reduce the parameter size, we always compress all of the projections within a block with the same block-dependent compression ratio. However, unlike other approaches~\cite{cai2025fastflux, PPCL}, we explicitly do not remove or substitute the whole block through a linear layer~\cite{cai2025fastflux}, in order to retain the nonlinear interactions within it.

For model compression, we propose to use truncated SVD, removing the least dominant singular values. Let $W \in \mathbb{R}^{n\times m}$ be a linear projection matrix. We compute its SVD and retain only the top $r$ singular values and vectors for our truncated approximation as
    \begin{align}
        W = U \Sigma V^T \approx  U_r \Sigma_r V^T_r = BA,
    \end{align}
where $B = U_r \sqrt{\Sigma_r} \in \mathbb{R}^{n\times r}$  and  $A = \sqrt{\Sigma_r}V^T_r \in \mathbb{R}^{r\times m}$ for numerical stability. With the original parameter count $P_0 = mn$ and the truncated parameter count $P_r = r(m + n)$, the compression ratio is $c = 1 - P_r / P_0$. Imposing a target compression ratio $c \in (0, 1)$ gives $r(n + m) \leq (1-c)nm$, which determines the rank for the truncation as
    \begin{align}
        r \leq \frac{(1 - c)nm}{n + m}.
    \end{align}
In practice, $r$ is chosen as the largest integer satisfying this constraint. Since not all transformer blocks contribute equally to the final model quality\cite{cnxs, HierarchPrune, deltadit}, we compress only a subset of blocks. Let the model consist of blocks indexed by $i \in \{1,\dots,N\}$ and let $\mathbb{S} \subseteq \{1,\dots,N\}$ denote the subset of blocks selected for compression according to the importance ranking described in \cref{subsec:method_block_analysis}.
Each block $i$ contains projection matrices with total parameter count $P_0^i$. For compressing a block $i \in \mathbb{S}$, we apply truncated SVD with a block-specific compression ratio $c_i \in (0,1)$, while all blocks $i \notin \mathbb{S}$ remain unchanged.

For a compressed block $i \in \mathbb{S}$, the parameter count changes after compression as $P_0^i \longrightarrow (1 - c_i) P_0^i$. For the block-specific rank-assignment, let the compressed blocks be ordered according to increasing importance. We assign compression ratios using a linearly decreasing schedule
    \begin{align}
        c_i = c_m - \alpha\  \text{order}(i),
        \label{eq:ranked_compression}
    \end{align}
with $c_m$ being the largest compression of a block and  $\alpha > 0$ the slope parameter. Blocks with higher importance receive less compression (\ie smaller $c_i$). We choose the slope $\alpha$ such that we achieve the desired global parameter budget 
    \begin{align}
        P \coloneq \frac{\sum_{i\notin \mathbb{S}} P_0^i + \sum_{j \in \mathbb{S}} (1 - c_j) P_0^j}{\sum_k P_0^k}.
        \label{eq:global_compression}
    \end{align}
See \Cref{fig:method} (right) for an illustration. By compressing only a subset of all blocks, we maintain potentially critical blocks unaltered and only focus on compressing blocks with the highest expected redundancy.

\subsection{Fine-Tuning}
\label{subsec:method_training_and_loss}

We now fine-tune the compressed student model $f_\theta$ with two complementary objectives in mind: \emph{(1)} preserving task performance for text-to-image generation, and \emph{(2)} transferring knowledge from the uncompressed teacher $f$.

\textbf{Rectified Flow Objective.}
To preserve generative performance, we train the student with the same rectified-flow objective as the teacher. Following FLUX, we train the student with the rectified flow velocity prediction objective. Let $x\sim p_\text{data}, \epsilon \sim \mathcal{N}(0, I)$, and $t\sim \mathcal{U}(0, 1)$. We define the linear interpolation $z_t = (1 - t) x + t\epsilon$ with target velocity $v^\ast = \epsilon - x$. The student model predicts a velocity field $v_\theta (z_t, t, y)$ conditioned on text $y$ and we optimise
    \begin{align}
    	\mathcal{L}_\text{RF} = \mathbb{E}_{x,\epsilon,t,y}\Big[ \big\Vert v_\theta (z_t, t, y) - (\epsilon - x) \big\Vert_2^2 \Big].
    \end{align}

\textbf{Modular Knowledge Distillation.}
To leverage the uncompressed model’s knowledge, we additionally employ knowledge distillation. Instead of simply calculating the loss on the final generated images \cite{salimans2022progressive, ADD}, we use a more complete variant. In contrast to substituting or pruning of whole blocks, we apply redundancy-ranked compression, which preserves the block structure of the original model. This enables us to perform module-wise alignment of intermediate representations, similar to \cite{KDD}. The modular knowledge distillation (MKD) loss combines velocity-level distillation
    \begin{align}
    	\mathcal{L}_\text{KD} = \mathbb{E}_{x, \epsilon, t, y} \Big[ \big\Vert v_\text{T}(z_t, t, y) - v_\theta(z_t, t, y) \big\Vert_2^2 \Big]
    \end{align}
where $v_\text{T}(z_t, t, y)$ is the prediction of the teacher, with module-wise feature distillation. Let $f^i$ and $f^i_\theta$ denote the $i$-th transformer block of the uncompressed teacher and the compressed student, respectively, and let 
    \begin{align}
    	h_i = f^i(h_{i-1}; t, y), \qquad \hat h_i = f_\theta^i(\hat h_{i-1}; t, y), \qquad h_0 = \hat h_0 = \text{embed}(z_t, y),
    \end{align}
be the intermediate feature maps. We penalise deviations of intermediate features after each block with 
    \begin{align}
    	\mathcal{L}_\text{F} = \sum_i \lambda_i \mathbb{E}_{x, \epsilon, t, y}\left[ \text{MSE}(h_i, \hat h_i) \right].
    \end{align}  
We set $\lambda_i$ to equalise the magnitude of block-wise contributions. The MKD loss is then given as
    \begin{align}
    	\mathcal{L}_\text{MKD} &= \lambda_\text{F}\mathcal{L}_\text{F} + \lambda_\text{KD} \mathcal{L}_\text{KD},
    \end{align}
with the complete fine-tuning objective being
    \begin{align}
    	\mathcal{L} = \lambda_\text{RF}\mathcal{L}_\text{RF} + \mathcal{L}_\text{MKD} = \lambda_\text{RF}\mathcal{L}_\text{RF} + \lambda_\text{F}\mathcal{L}_\text{F} + \lambda_\text{KD}\mathcal{L}_\text{KD}.
        \label{eq:final_loss}
    \end{align}

\section{Experiments}
\label{sec:experiments}

    \begin{figure}[tb]
      \centering
        \centering
        \includegraphics[width=.7\textwidth]{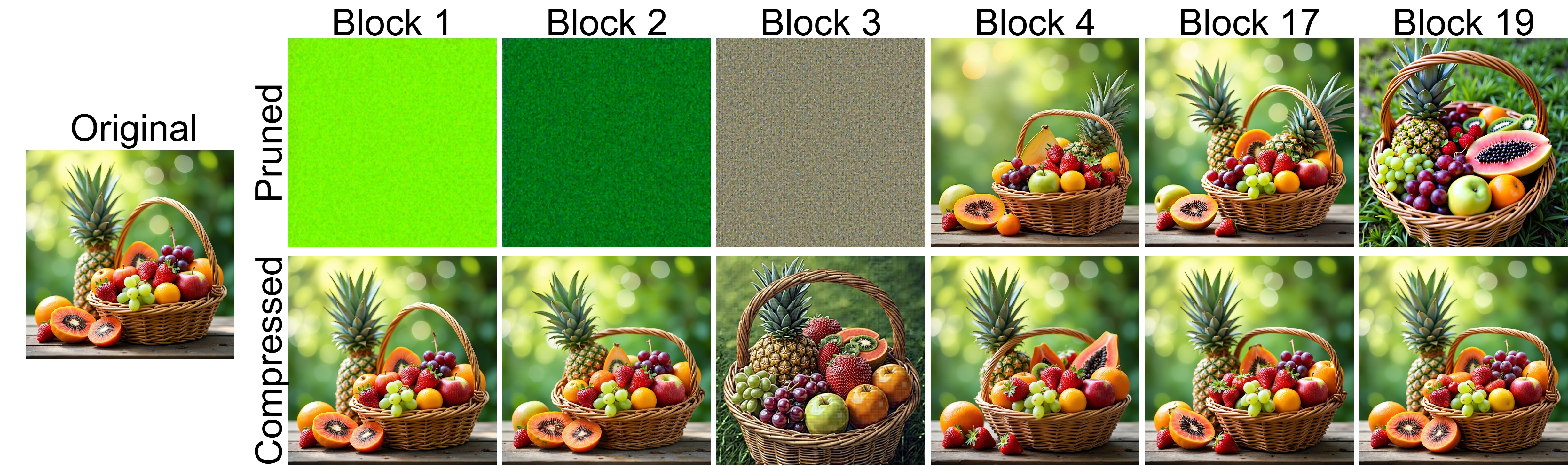}
      \caption{\textbf{Block sensitivity comparison}.  Generated images after probing double-stream blocks of FLUX.dev\cite{flux2024} with pruning \textit{(top)} \vs truncated SVD \textit{(bottom)}.}
      \label{fig:method_pruning_visuals}
    \end{figure}

We first provide the experimental setup in \cref{subsec:exp_setup}. Before presenting the main comparison to the state of the art in \cref{subsec:exp_trained_comparison}, we present two experiments that motivate our overall methodology. 
First, in \cref{subsec:exp_pruning_vs_compression} we validate that redundancy in DiTs is distributed across projection matrices throughout the whole network, rather than at the level of entire Transformer blocks. Second, in \cref{subsec:exp_training_free_compression} we show that purely applying truncated SVD to DiTs without re-training does not result in catastrophic degradation as for LLMs. The section concludes with an application to a temporally-distilled model in \cref{subsec:exp_distilled_models} and an ablation study in \cref{subsec:exp_ablations}.

\subsection{Experimental Setup}
\label{subsec:exp_setup}

\textbf{Generative Model.} We compress FLUX.dev\cite{flux2024} using our \ours approach. 
This preserves the original block structure and non-linearities. The VAE and text-encoder are kept frozen and are not included in reported parameter budgets. We report the remaining parameter budget $P$ with respect to the DiT backbone parameters for the Diffusion Transformer weights only.

\textbf{Importance Analysis.} To estimate block importance for our final models (\cref{subsec:method_block_analysis}), we sample $K_{\text{probe}}=
\num{1000}$ images from LAION\cite{LAION} and generate captions using JoyCaption \cite{joycaption}. For each candidate block $i$, we construct a probed model by compressing only block $i$ with a fixed block compression ratio of $c_m =\num{0.6}$ while keeping all other blocks uncompressed, and generate paired image sets using identical prompts.

\textbf{Importance metric.} We quantify degradation using CMMD\cite{CMMD} between the probed and uncompressed image sets, and rank blocks by their induced degradation. Lower degradation suggests higher redundancy, hence we perform stronger compression.

\textbf{Fine-tuning Settings.} For distillation, we train on $K_{\text{train}}=\num{190000}$ images generated by FLUX.dev conditioned on captions from JoyCaption on LAION images. We optimise the student using the combined objective in \cref{eq:final_loss}. Fine-tuning is done for \qty{40}{\kilo\count} steps with a batch size of 4 using Adafactor~\cite{Adafactor} with learning rate \num{1.8e-6} on one H200 GPU. This is used to achieve a parameter budget of \qty{68}{\percent}, as in alternative approaches\cite{PPCL, flux_lite, chromaHD}. To compress further, we re-evaluate the block importance and train subsequently for \qty{20}{\kilo\count} more steps.

\textbf{Compression configuration.} Unless stated otherwise, we compress $|\mathbb{S}|=\num{30}$ blocks using variable maximum block-compression ratios $c_m$, which we specify in the supplement, and determine the linear schedule slope $\alpha$ to satisfy the target global budget P (Eq.\ \ref{eq:global_compression}).

\textbf{Evaluation.} We report results on three public benchmarks: GenEval\cite{GenEval}, which evaluates object-centric compositional text–image alignment; DPG\cite{DPG}, which measures faithfulness to dense prompts; and HPSv2\cite{HPSv2}, which combines prompt–image alignment with approximated human preference. All metrics are evaluated using the same inference protocol, \ie a resolution of $\num{1024}\times \num{1024}$, \num{50} steps, classifier-free guidance scale \num{3.5}, and fixed seeds per prompt. We additionally report peak VRAM usage with bf16 precision under the same settings. For clarity, we report the average quality reduction across those benchmarks in relation to the respective uncompressed model as $R$.

\subsection{Block Sensitivity Probing}
\label{subsec:exp_pruning_vs_compression}

We first investigate how Diffusion Transformers react to structural changes at the block level. For each block $i$, we consider two variants: \emph{(1)} Pruning, where the entire block is replaced by an identity mapping, similarly to \cite{bk-sdm}, and \emph{(2)} single-block compression, where block $i$ is compressed using truncated SVD with fixed block-compression ratio $c_m=0.6$, while all other blocks remain unchanged. Furthermore, the blocks are not fine-tuned at this stage. 
For each probed model, we generate images using the same prompts as the uncompressed model. We quantify degradation using CMMD between the generated image sets, while the CMMD score is normalised by the number of reduced parameters.

\textbf{Pruning \vs Compression.}
\cref{fig:method_pruning_visuals} shows that pruning frequently leads to severe degradation and, for early blocks, catastrophic failure of the generation process. In contrast, single-block compression results in substantially smaller degradation across all blocks. Quantitatively, pruning yields a mean CMMD of $1.18 \pm 3.36$ per probed block, whereas block compression achieves $0.23 \pm 0.65$. This indicates that rank reduction preserves functional behaviour far more gracefully than removing entire blocks. In \cref{subsec:exp_training_free_compression}, this trend is further confirmed by comparing our compression schedule with pruning under equal parameter budgets in the training-free setting. These results suggest that redundancy is distributed across projection matrices within blocks throughout the network, while complete block removal disrupts essential transformations. We use this setup to derive the block-importance ranking used in \cref{eq:ranked_compression}.

\subsection{Training-Free Model Compression}
\label{subsec:exp_training_free_compression}

\begin{table}[t]
\centering
{%
\tinyscript
\begin{tabularx}{0.7\textwidth}{@{}XS[table-format=3]S[table-format=1.3]S[table-format=2.2]S[table-format=2.1]S[table-format=2.2]@{}}
\toprule
\textbf{Method} & {\textbf{P}[\%] $\downarrow$} & \textbf{GenEval} $\uparrow$ & \textbf{HPSv2} $\uparrow$ & \textbf{DPG} $\uparrow$ & \textbf{R}[\%]$\downarrow$ \\
\midrule
\textcolor{gray}{FLUX.dev~\cite{flux2024}}
    & \color{gray} 100
    & \color{gray} 0.647
    & \color{gray} 31.70 
    & \color{gray} 83.9 
    & \color{gray} 0.00 
    \\
\midrule
EcoDiff~\cite{EcoDiff} &
    90 &
    0.452 &
    29.67 &
    78.0 &
    14.52 
    \\
Pruning &
    90 &
    \bfseries 0.646 &
    \bfseries 31.02 &
    \bfseries 83.3 &
    \bfseries 1.00 
    \\
\textbf{\ours (ours)} &
    90 &
    0.638 &
    29.03 &
    78.9 &
    5.24
    \\
\midrule
EcoDiff~\cite{EcoDiff} &
    80 &
    0.187 &
    23.63 &
    49.1 &
    45.99 
    \\
Pruning &
    80 &
    0.381 &
    24.88 &
    65.2 &
    28.31 
    \\
\textbf{\ours (ours)} &
    80 &
    \bfseries 0.617 &
    \bfseries 28.86 &
    \bfseries 78.4 &
    \bfseries 6.69 
    \\
\midrule
EcoDiff~\cite{EcoDiff}&
    70 &
    0.002 &
    7.26 &
    17.0 &
    85.50 
    \\
Pruning &
    70 &
    0.000 &
    4.37 &
    2.2 &
    94.52 
    \\
\textbf{\ours (ours)} &
    70 &
    \bfseries 0.580 &
    \bfseries 28.48 &
    \bfseries 76.6 &
    \bfseries 9.86
    \\
\bottomrule
\end{tabularx}
}
\smallskip
\caption{\textbf{Training-free compression.} While importance-aware Pruning and EcoDiff collapse even under moderate compression, our \ours maintains stable generation quality even at 70\% of the original parameter count.}
\label{tab:exp_training_free}
\end{table}
To evaluate the intrinsic behaviour of Diffusion Transformers under low-rank approximation, we evaluate our importance-aware SVD compression without any subsequent fine-tuning. In this setting, truncated SVD is applied once according to the importance-guided schedule from \cref{eq:ranked_compression}, and the resulting model is evaluated directly.
The results are reported in \cref{tab:exp_training_free}. At 90\,\% parameter count, our \ours degrades only 5.24\,\% in average performance, compared to 14.52\,\% for EcoDiff, a state-of-the-art training-free pruning method. At 80\,\%, \ours degrades by 6.69\,\%, whereas EcoDiff drops by 45.99\,\%. At approximately 70\,\% parameters, EcoDiff collapses almost entirely with $R=85.50\,\%$, while our method drops only by $R=9.86\,\%$.

These results indicate that low-rank compression on projection-matrix level preserves the functional structure of Diffusion Transformers to a surprising extent, even without fine-tuning, while leading to potential catastrophic failure in LLMs\cite{svdllm, FWSVD}. In contrast, pruning-based removal of blocks\cite{bk-sdm} leads to high performance at minimal parameter reduction but rapidly collapses ($R=94.52\,\%$) under a moderate parameter budget of 70\,\%. While fine-tuning further improves alignment with the uncompressed teacher (\cref{subsec:exp_trained_comparison}), these findings show that the effectiveness of our approach is not solely driven by fine-tuning, but reflects an inherent tolerance of Diffusion Transformers to structured rank reduction.

\subsection{Comparison to State-of-the-Art}
\label{subsec:exp_trained_comparison}
\begin{table}[t]
\centering
{%
\tinyscript
\begin{tabularx}{\textwidth}{@{}XlS[table-format=3]S[table-format=3.1]S[table-format=1.3]S[table-format=2.1]S[table-format=2.1]S[table-format=2.2]S[table-format=2]@{}}
\toprule
      \bfseries Method
    & \bfseries Venue 
    & {{\bfseries Params.\ }[\%] $\downarrow$} 
    & {{\bfseries VRAM} [\%] $\downarrow$} 
    & {{\bfseries GenEval }$\uparrow$} 
    & {{\bfseries HPSv2 }$\uparrow$} 
    & {{\bfseries DPG }$\uparrow$} 
    & {{\bfseries R }[\%]$\downarrow$} 
    & {{\bfseries Rank }$\downarrow$}
    \\
\midrule
\textcolor{gray}{FLUX.dev~\cite{flux2024}}
    & \color{gray} {--} 
    & \color{gray} 100
    & \color{gray} 100.0
    & \color{gray} 0.647
    & \color{gray} 31.70 
    & \color{gray} 83.9 
    & \color{gray} 0.00 
    & \color{gray} {--} 
    \\
\midrule
Chroma-HD~\cite{chromaHD}
    & HF 
    & 75 
    & 82.5 
    & 0.593 
    & {--}
    & \bfseries 84.0 
    & 4.09
    & 3
    \\
FluxLite~\cite{flux_lite}
    & HF'24 
    & 68 
    & 78.8 
    & 0.523 
    & \bfseries 31.31 
    & 79.3 
    & 8.61 
    & 5 
    \\
TinyFusion~\cite{tinyFusion}
    & CVPR'25 
    & 68 
    & 74.4 
    & 0.511
    & {--}
    & 77.2 
    & 14.48 
    & 6 
    \\
Dense2MoE~\cite{dense2Meo}
    & ICCV'25 
    & 75
    & {--}
    & 0.403
    & {--}
    & 73.6
    & 24.97
    & 9 
    \\
FastFlux~\cite{cai2025fastflux}
    & AAAI'26
    & 80 
    & {--}
    & 0.530 
    & 27.26 
    & {--}
    & 16.04 
    & 8 
    \\
PPCL~\cite{PPCL}
    & CVPR'26
    & 68 
    & 74.4 
    & 0.605 
    & {--}
    & 80.0
    & 5.55
    & 4
    \\
Hier.Prune~\cite{HierarchPrune}
    & AAAI'26 
    & 68 
    & 74.4 
    & 0.503 
    & {--}
    & 75.7
    & 15.99 
    & 7
    \\
EcoDiff~\cite{EcoDiff}
    & ICLR'26
    & 80 
    & {--}
    & 0.399 
    & 25.99 
    & {--}
    & 28.17 
    & 10 
    \\
\textbf{\ours-m} &
    {--} &
    68 &
    78.2 &
    \bfseries 0.645 &
    31.27 &
    83.4 &
    \bfseries 0.75 &
    1 
    \\
\textbf{\ours-s} &
    {--} &
    \bfseries 57 &
    \bfseries 70.7 &
    0.616 &
    31.22 &
    82.6 &
    2.60 &
    2 
    \\
\bottomrule
\end{tabularx}
}
\smallskip
\caption{\textbf{Quantitative results} for compression of FLUX.dev~\cite{flux2024}. The final Rank is based on the average quality reduction $R$. Our medium-sized model \ours-m achieves first rank. Notably, our small-sized model \ours-s, which has considerably fewer parameters than all alternative approaches, is runner-up.}
\label{tab:exp_quantitative}
\end{table}
\begin{table}[t]
\centering
{%
\tinyscript
\begin{tabularx}{.8\textwidth}{@{}XcS[table-format=3]S[table-format=3.1]S[table-format=1.3]S[table-format=2.1]S[table-format=2.1]S[table-format=2.2]S[table-format=2]@{}}
\toprule
      \bfseries Method
    & {{\bfseries Params.\ }[\%] $\downarrow$} 
    & {{\bfseries GenEval }$\uparrow$} 
    & {{\bfseries HPSv2 }$\uparrow$} 
    & {{\bfseries DPG }$\uparrow$} 
    & {{\bfseries R }[\%] $\downarrow$} 
    \\\midrule
\textcolor{gray}{PixArt-$\Sigma$\cite{pixart_sigma}} 
    & \color{gray} 100 (0.6B)
    & \color{gray} 0.535
    & \color{gray} 30.50 
    & \color{gray} 78.8
    & \color{gray} 0.00
    \\
\midrule
\textbf{\ours} &
    90 &
    0.543 &
    29.34 &
    78.5 &
    0.22 
    \\
\textbf{\ours} &
    80 &
    0.528 &
    29.79 &
    78.0 &
    1.62 
    \\
\textbf{\ours} &
    70 &
    0.532 &
    29.64 &
    78.2 &
    1.38 
    \\
\textbf{\ours} &
    60 &
    0.528 &
    29.34 &
    78.0 &
    2.04 
    \\
\textbf{\ours} &
    50 &
    0.496 &
    28.28 &
    76.6 &
    5.79 
    \\
\bottomrule
\end{tabularx}
}
\smallskip
\caption{\textbf{Results for PixArt-$\Sigma$ model.}
The degradation in quality of our \ours results are gradual, without abrupt failure.}
\label{tab:exp_pixart}
\end{table}

\textbf{Quantitative Results.}
We compare our approach with recent compression methods at matched parameter budgets after fine-tuning in \cref{tab:exp_quantitative}. Unless stated otherwise, models are re-evaluated under a unified inference protocol as described in \cref{subsec:exp_setup}. For TinyFusion \cite{tinyFusion}, Dense2MoE \cite{dense2Meo}, Chroma-HD \cite{chromaHD}, and HierarchicalPrune \cite{HierarchPrune}, we report numbers from PPCL \cite{PPCL} due to the absence of publicly available checkpoints or inference code.

At a parameter budget of 68\,\%, our model (\ours-m) achieves near full performance across GenEval, HPSv2, and DPG, with a degradation of only $R=0.75\,\%$ compared to FLUX.dev. Notably, this exceeds competing methods at the same (or even larger) parameter budgets, while preserving strong compositional reasoning on GenEval.
At a more aggressive 57\,\% parameter budget, our model (\ours-s) maintains competitive performance, with only $R=2.6\,\%$.
Despite operating at a substantially smaller parameter footprint than competing approaches at 68\,\%, it outperforms them on GenEval and remains competitive on HPSv2 and DPG.
These results establish a new quality--parameter frontier for compressed Diffusion Transformers, demonstrating that importance-aware low-rank distillation enables substantial parameter reduction without catastrophic degradation. Full quantitative tables are provided in the supplement.

\textbf{Memory Efficiency.}
In addition to parameter count, we report peak VRAM usage under identical inference settings. \ours achieves competitive or superior memory efficiency compared to pruning-based methods at comparable parameter budgets, highlighting the practical benefits of truncated SVD.

\textbf{Progressive Compression.}
To analyse the trade-off between parameter budget and generation quality, we evaluate models across a range of global parameter budgets $P \in [0.9, 0.4]$. The results are shown in \cref{fig:front_figure_plot}.
For mild compression (\eg, $P \geq 0.68$), performance remains nearly indistinguishable from the uncompressed model across all evaluated metrics, with  $<1\,\%$ quality reduction on average. As compression increases, quality degrades gradually and predictably. Notably, substantial performance deterioration only becomes apparent beyond approximately 40–50\,\% parameter reduction.
This indicates that DiTs contain significant structured redundancy that can be removed without immediate degradation. However, beyond a certain compression level, the reduction in representational capacity begins to affect generation quality more noticeably.
Since FLUX.1-dev contains 12B parameters, its robustness to compression could partly stem from overparameterization. To test whether this behavior extends beyond large-scale models, we additionally evaluate SVDtrunc on PixArt-$\Sigma$ \cite{pixart_sigma} in \cref{tab:exp_pixart}. PixArt-$\Sigma$ shows the same quantitative trend of graceful degradation under increasing compression, suggesting that the redundancy exploited by SVDtrunc is not specific to FLUX.1-dev but is present across pretrained DiTs of different sizes and architectures.

\begin{figure}[tb]
        \centering
        \includegraphics[width=1.\linewidth]{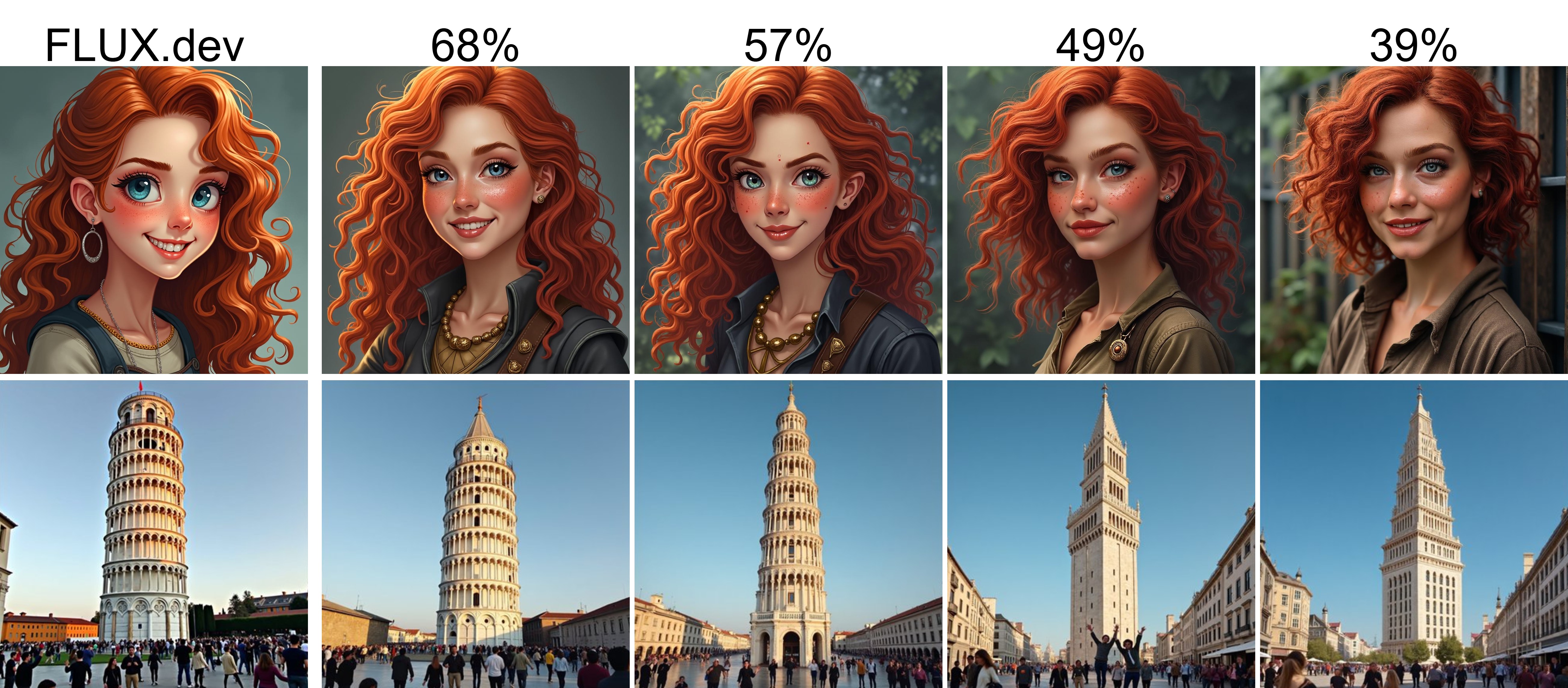}
        \caption{Example of domain drift (top) and memorisation loss (bottom) under different parameter budgets.}
        \label{fig:domain_drift}
    \end{figure}

\textbf{Qualitative Results and Domain Drifts.}
In all our tests, our compressed models remain competitive with, or outperform, existing compression approaches. Qualitative examples are shown in \cref{fig:front_figure_images}. In visual inspection, we noticed an occasional domain drift, predominatly at stronger compression levels where the generated images move towards slightly different stylistic domains while remaining visually coherent. \cref{fig:domain_drift} illustrates such a domain drift from cartoonish style to realistic portrait.
Importantly, this stylistic drift is not consistently accompanied by a measurable degradation in visual quality and benchmark scores remain relatively stable.
We conjecture that this behaviour indicates that \ours preserves the parameters most critical for generation quality, while stylistic characteristics are more susceptible to changes under compression. As a result, the model retains its core generative capabilities even under substantial parameter reduction.

Beyond the stylistic domain drift, decreasing the parameter budget can reduce the model's ability to reproduce specific characters or highly distinctive concepts. In \cref{fig:domain_drift}, we show examples illustrating this behaviour. Well-known instances such as the tower of Pisa (\cref{fig:domain_drift}) or Bugs Bunny, Darth Vader, or the Minions (supplement) are no longer reproduced faithfully at stronger compression levels. Instead, the model generates high-quality images resembling the general concept but lacking the distinctive identity of the original entity.

This behaviour suggests that part of the model capacity is used to memorise specific instances or highly detailed concepts, whereas general visual quality can be maintained with substantially fewer parameters. Intuitively, generating plausible images primarily requires learning reusable visual features and compositional patterns, whereas representing highly specific identities or characters requires additional capacity. Consequently, under strong compression, models may begin to discard parameters associated with memorised instances while retaining those necessary for general image synthesis. This observation indicates that substantial parameter redundancy in large diffusion transformers may be related to storing highly specific concepts rather than the core mechanisms required for image generation. We refer to the supplement for further examples.

\subsection{Application to Temporally-Distilled Model}
\label{subsec:exp_distilled_models}

\begin{table}[t]
\centering
{%
\tinyscript
\begin{tabularx}{0.7\textwidth}{@{}XS[table-format=3]S[table-format=1.3]S[table-format=2.2]S[table-format=2.1]S[table-format=2.2]@{}}
\toprule
\textbf{Method} 
& {\textbf{P}[\%] $\downarrow$} 
& \textbf{GenEval} $\uparrow$ 
& \textbf{HPSv2} $\uparrow$ 
& \textbf{DPG} $\uparrow$ 
& \textbf{R}[\%]$\downarrow$
\\
\midrule
\textcolor{gray}{FLUX.schnell\cite{flux2024}} 
    & \color{gray} 100 
    & \color{gray} 0.667 
    & \color{gray} 30.85 
    & \color{gray} 85.0 
    & \color{gray} 0.00
    \\
\midrule
\textbf{\ours} (ours) &
    90 &
    0.666 &
    29.59 &
    85.2 &
    1.34 
    \\
\textbf{\ours} (ours) &
    80 &
    0.668 &
    29.48 &
    84.9 &
    1.48 
    \\
\textbf{\ours} (ours) &
    70 &
    0.648 &
    29.30 &
    84.8 &
    2.72 
    \\
\textbf{\ours} (ours) &
    60 &
    0.621 &
    29.05 &
    84.3 &
    4.53 
    \\
\bottomrule
\end{tabularx}
}
\smallskip
\caption{\textbf{Results for temporally-distilled FLUX.schnell \cite{flux2024} model.}
The degradation in quality of our \ours results are gradual, without abrupt failure.}
\label{tab:exp_distilled}
\end{table}
Step distillation reduces the number of denoising steps, compressing the generation process along the temporal dimension \cite{ADD, salimans2022progressive, sdxl-lightning}. One might therefore expect temporally distilled models to exhibit reduced internal redundancy and increased sensitivity to parameter compression. We test this hypothesis on the official temporally-distilled model FLUX.schnell~\cite{flux2024}, which generates images in four steps.
We recompute the block-importance ranking and apply projection-level low-rank compression using the same importance-aware allocation strategy from \cref{subsec:method_block_analysis}. Results are summarised in \cref{tab:exp_distilled}.
At a 90\,\% parameter budget, GenEval performance remains virtually unchanged and DPG stays comparable. Even at 80\,\%, the compressed model maintains near-baseline performance across all metrics. As compression increases to 70\,\% and 60\,\%, degradation remains gradual and controlled, without abrupt failure. However, the degradation is slightly higher, with $R=4.53\,\%$ at $P=60\,\%$, compared to FLUX.dev with $R=2.60\,\%$ at $P=57\,\%$ (see \cref{tab:exp_quantitative}).
These results indicate that tolerance to projection-level low-rank compression persists even after temporal distillation. Redundancy therefore appears to be structurally embedded in the Transformer backbone rather than arising solely from multi-step diffusion dynamics.

\subsection{Ablations}
\label{subsec:exp_ablations}

We ablate the rank-allocation strategy, which is the central design choice of \ours. This experiment analyses how importance-aware rank allocation affects generation quality under decreasing parameter budgets. Additional ablations on the distribution of compression across blocks and the fine-tuning objective are provided in the supplement.

\textbf{Importance-Aware \vs Uniform Allocation.}
\begin{table}[t]
\centering
{%
\tinyscript
\begin{tabularx}{0.6\textwidth}{@{}XS[table-format=3]S[table-format=1.3]S[table-format=2.2]S[table-format=2.1]S[table-format=2.2]@{}}
\toprule
\textbf{Method} & {\textbf{P}[\%] $\downarrow$} & \textbf{GenEval} $\uparrow$ & \textbf{HPSv2} $\uparrow$ & \textbf{DPG} $\uparrow$ & \textbf{R}[\%]$\downarrow$ \\
\midrule
\textcolor{gray}{FLUX.dev} 
    & \textcolor{gray}{100} 
    & \textcolor{gray}{0.647} 
    & \textcolor{gray}{31.70} 
    & \textcolor{gray}{83.9} 
    & \textcolor{gray}{0.00} 
    \\
\midrule
Random &
    68 &
    0.633 &
    31.19 &
    82.9 &
    1.63 
    \\
Uniform &
    68 &
    0.633 &
    \textbf{31.38} &
    82.8 &
    1.46 
    \\
\textbf{Ours} &
    68 &
    \textbf{0.645} &
    31.27 &
    \textbf{83.4} &
    \textbf{0.75} 
    \\
\midrule
Random &
    57 &
    0.588 &
    30.22 &
    81.1 &
    5.68 
    \\
Uniform &
    57 &
    0.613 &
    30.92 &
    82.3 &
    3.19 
    \\
\textbf{Ours} &
    57 &
    \textbf{0.616} &
    \textbf{31.22} &
    \textbf{82.6} &
    \textbf{2.60} 
    \\
\midrule
Random &
    49 &
    0.542 &
    29.16 &
    79.8 &
    9.69 
    \\
Uniform &
    49 &
    0.582 &
    29.96 &
    81.2 &
    6.23 
    \\
\textbf{Ours} &
    49 &
    \textbf{0.587} &
    \textbf{31.01} &
    \textbf{81.7} &
    \textbf{4.68} 
    \\
\bottomrule
\end{tabularx}
}
\caption{\textbf{Compression Schedule Ablation.} We evaluate uniform, random and importance-aware (our) SVD compression schedules after training.}
\label{tab:abl_uniform_vs_ours}
\end{table}
\cref{tab:abl_uniform_vs_ours} compares uniform, random block, and importance-aware compression across multiple parameter budgets. With uniform, we compress all blocks equally, while the random schedule has a random ranking order for \cref{eq:ranked_compression}. All three strategies exhibit smooth degradation as compression increases, indicating that Diffusion Transformers tolerate projection-level rank reduction even under naive allocation.
At moderate compression (68\,\%), performance differences remain small, with \ours retaining near-teacher performance. As compression becomes more aggressive (57\,\% and 49\,\%), importance-aware allocation consistently outperforms both uniform and random schedules across GenEval, HPSv2, and DPG, yielding improved quality retention and lower relative degradation.
These results indicate that redundancy is broadly distributed across the network, enabling stable low-rank compression even under uniform allocation, while importance-aware rank allocation becomes increasingly beneficial under stronger constraints.

\section{Summary and Conclusion}
\label{sec:conclusion}

We investigated truncated SVD for compressing large-scale Diffusion Transformers and find that, unlike autoregressive LLMs, DiTs exhibit substantial tolerance to low-rank approximation on the projection-matrix level. Even under significant parameter reduction, generative performance degrades gradually rather than catastrophically, and this behaviour is already observable in a training-free setting.
Across benchmarks, our approach achieves a favourable quality--parameter trade-off and remains robust even when applied to models after temporal distillation, positioning low-rank compression on the projection-matrix level as a promising direction for further efficiency gains. At aggressive compression levels, domain drift can occur, highlighting limits of global representational capacity and motivating future work on preserving stylistic diversity under stronger parameter constraints.

\section*{Acknowledgments}
Denis Zavadski and Sebastian Heid are supported by the Konrad Zuse School of Excellence in Learning and Intelligent Systems (\href{https://eliza.school/}{ELIZA}) through the DAAD programme Konrad Zuse Schools of Excellence in Artificial Intelligence, sponsored by the German Federal Ministry of Education and Research. The authors gratefully acknowledge the support by the Ministry of Science, Research and the Arts Baden-Württemberg (MWK) through bwHPC, SDS@hd and the German Research Foundation (DFG) through the grants INST 35/1597-1 FUGG and INST 35/1503-1 FUGG. This project has received funding from the European Research Council (ERC) under the European Union’s Horizon 2020 research and innovation programme (grant agreement No.\ 866008), and from the DFG under Germany’s Excellence Strategy (EXC-3057/1 “Reasonable Artificial Intelligence”, Project No.\ 533677015). We thank Tim Küchler for the discussions.

{
    \small
    \bibliographystyle{splncs04.bst}
    \bibliography{main}
}

\newpage
\appendix
\setcounter{figure}{0}
\setcounter{table}{0}

\renewcommand{\thefigure}{S\arabic{figure}}
\renewcommand{\thetable}{S\arabic{table}}

\begin{figure}
    \centering
    \includegraphics[width=.9\linewidth]{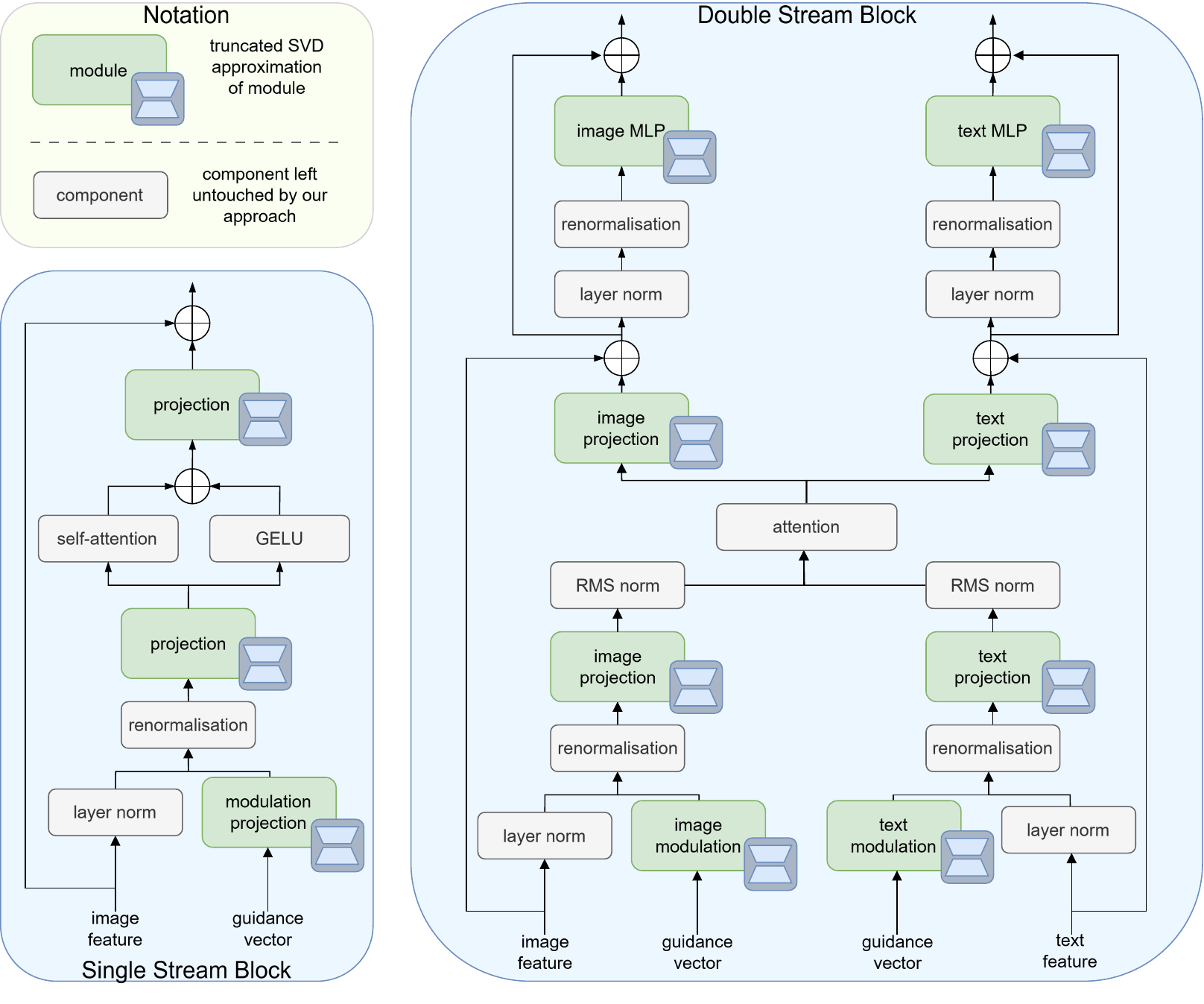}
    \caption{\textbf{Conceptual SVD Compression.} High-level illustration of where truncated SVD–based low-rank approximations are applied within multi-modal diffusion transformer (MMDiT) blocks \cite{sd3_mmdit}. Green modules denote components compressed with block-wise low-rank approximations, while grey components contain negligible or no parameters and are left unchanged to preserve the original architecture.}
    \label{fig:sup_MMDiT}
\end{figure}

This supplement provides additional details and analyses supporting the main paper. We first illustrate the block-wise SVD compression applied to multi-modal diffusion transformer blocks in \cref{sup:block_wise_compression}. \cref{sup:add_quant_results} presents additional quantitative results across multiple compression levels on GenEval\cite{GenEval}, DPG\cite{DPG}, and HPSv2\cite{HPSv2}. Training hyperparameters and compression schedules are reported in \cref{sup:training_specs} while \cref{sup:ablations} provides additional ablations of the distribution of compression across blocks, and the choice of fine-tuning objective. Finally, we provide further examples for qualitative effects that arise under stronger compression, including stylistic domain drift and reduced memorisation of specific concepts (\cref{sup:domain_drift}) as well as additional qualitative examples (\cref{sup:add_qual_examples}).

\section{Block-Wise SVD Compression}
\label{sup:block_wise_compression}

We illustrate the high-level composition of multi-modal diffusion transformer blocks in \cref{fig:sup_MMDiT}. The majority of block parameters are located in the projection modules (highlighted green). The remaining components (grey) have rather no or negligible parameter counts, so we leave them untouched. By compressing projections rather than removing or replacing entire blocks, the internal transformations and interactions within each block are preserved and the overall network architecture remains unchanged. All modules within one transformer block are compressed with a block-specific compression ratio $c$.

\section{Additional Quantitative Results}
\label{sup:add_quant_results}
\begin{table}[t]
\centering
{%
\tinyscript
\begin{tabularx}{.8\textwidth}{@{}XlS[table-format=3]S[table-format=3.1]S[table-format=1.3]S[table-format=2.1]S[table-format=2.1]S[table-format=2.2]S[table-format=2]@{}}
\toprule
      \bfseries Method
    & {{\bfseries Params.\ }[\%] $\downarrow$} 
    & {{\bfseries GenEval }$\uparrow$} 
    & {{\bfseries HPSv2 }$\uparrow$} 
    & {{\bfseries DPG }$\uparrow$} 
    & {{\bfseries R }[\%] $\downarrow$} 
    \\
\midrule
\textcolor{gray}{FLUX.dev~\cite{flux2024}}
    & \color{gray} 100 (12B)
    & \color{gray} 0.647
    & \color{gray} 31.70 
    & \color{gray} 83.9 
    & \color{gray} 0.00 
    \\
\midrule
PPCL~\cite{PPCL}
    & 68 
    & 0.605 
    & {--}
    & 80.0
    & 5.55
    \\
\textbf{\ours} &
    90 &
    0.660 &
    31.36 &
    83.6 &
    -0.19 &
    \\
\textbf{\ours} &
    80 &
    0.651 &
    31.39 &
    83.6 &
    0.21 &
    \\
\textbf{\ours-m} &
    68 &
    0.645 &
    31.27 &
    83.4 &
    0.75 &
    \\
\textbf{\ours-s} &
    57 &
    0.616 &
    31.22 &
    82.6 &
    2.60 &
    \\
\textbf{\ours} &
    49 &
    0.587 &
    31.01 &
    81.7 &
    4.68 &
    \\
\textbf{\ours} &
    43 &
    0.545 &
    30.02 &
    79.9 &
    8.60 &
    \\
\textbf{\ours} &
    39 &
    0.495 &
    29.49 &
    78.0 &
    12.50 &
    \\
\midrule
\textcolor{gray}{PixArt-$\Sigma$\cite{pixart_sigma}} 
    & \color{gray} 100 (0.6B)
    & \color{gray} 0.535
    & \color{gray} 30.50 
    & \color{gray} 78.8
    & \color{gray} 0.00
    \\
\midrule
\textbf{\ours} &
    90 &
    0.543 &
    29.34 &
    78.5 &
    0.22 
    \\
\textbf{\ours} &
    80 &
    0.528 &
    29.79 &
    78.0 &
    1.62 
    \\
\textbf{\ours} &
    70 &
    0.532 &
    29.64 &
    78.2 &
    1.38 
    \\
\textbf{\ours} &
    60 &
    0.528 &
    29.34 &
    78.0 &
    2.04 
    \\
\textbf{\ours} &
    50 &
    0.496 &
    28.28 &
    76.6 &
    5.79 
    \\
\bottomrule
\end{tabularx}
}
\smallskip
\caption{\textbf{Quantitative results} for compression of FLUX.dev~\cite{flux2024} and PixArt-$\Sigma$\cite{pixart_sigma}. For clarity, we report the average quality reduction across GenEval\cite{GenEval}, HPSv2\cite{HPSv2} and DPG\cite{DPG} benchmarks in relation to the respective uncompressed model as $R$. \ours preserves most of the original model performance even at parameter budgets of $\leq$60\,\%, exhibiting graceful degradation as compression increases.}
\label{tab:sup_quant_summary}
\end{table}
In \cref{tab:sup_quant_summary}, we summarise the results of \ours compression on FLUX.dev~\cite{flux2024} across a range of compression levels, from low to high parameter reduction. For comparison, we report results of the strongest competing method, PPCL~\cite{PPCL}, at the common 68\,\% parameter budget.

Low-rank factorization replaces a dense projection with two smaller projections, which may introduce overhead at mild compression levels. For the stronger compression schedules used by \ours, however, the reduced projection cost lowers the measured per-step generator runtime from 424 ms for FLUX.1-dev to 380 ms for \ours-m and 354 ms for \ours-s, corresponding to relative runtimes of 89.6\% and 83.4\%, respectively.

Since FLUX.dev is a large-scale model with 12\,B parameters, we additionally evaluate compression on PixArt-$\Sigma$~\cite{pixart_sigma}, a significantly smaller diffusion transformer with 0.6\,B parameters~\cite{DiT}. This experiment demonstrates that the redundancy exploited by our method is inherent to diffusion transformers and persists across different model scales.

We report results on three public benchmarks: GenEval\cite{GenEval}, which evaluates object-centric compositional text–image alignment; DPG\cite{DPG}, which measures faithfulness to dense prompts; and HPSv2\cite{HPSv2}, which combines prompt–image alignment with approximated human preference.

\begin{table}[t]
\centering
{%
\tinyscript
\begin{tabularx}{.95\textwidth}{@{}XS[table-format=3]S[table-format=1.3]S[table-format=1.3]S[table-format=1.3]S[table-format=1.3]S[table-format=1.3]S[table-format=1.3]S[table-format=1.3]S[table-format=1.3]@{}}
\toprule
\textbf{Method}&
\multicolumn{1}{c}{Params. [\%]}  $\downarrow$&
\multicolumn{1}{c}{\makecell{Two\\Object}}  $\uparrow$&
\multicolumn{1}{c}{\makecell{Single\\Object}}  $\uparrow$&
\multicolumn{1}{c}{\makecell{Colour\\Attribute}}  $\uparrow$&
\multicolumn{1}{c}{\makecell{Colours}}  $\uparrow$&
\multicolumn{1}{c}{Counting}  $\uparrow$&
\multicolumn{1}{c}{Position}  $\uparrow$&
\multicolumn{1}{c}{Overall} $\uparrow$\\
\midrule
\textcolor{gray}{FLUX.dev\cite{flux2024}} 
    & \color{gray} 100  
    & \color{gray} 0.768  
    & \color{gray} 0.981  
    & \color{gray} 0.455  
    & \color{gray} 0.758  
    & \color{gray} 0.709  
    & \color{gray} 0.210  
    & \color{gray} 0.647  
    \\
\midrule
PPCL\cite{PPCL} &
    68 &    
    0.726 &    
    0.978 &    
    0.380 &    
    0.785 &    
    0.593 &    
    0.170 &    
    0.605 &    
    \\
\textbf{\ours} &
    90 &    
    0.806 &    
    0.984 &    
    0.473 &    
    0.769 &    
    0.703 &    
    0.223 &    
    0.656 &    
    \\
\textbf{\ours} &
    80 &    
    0.801 &    
    0.988 &    
    0.468 &    
    0.755 &    
    0.663 &    
    0.235 &    
    0.651 &    
    \\
\textbf{\ours-m} &
    68 &    
    0.818 &    
    0.991 &    
    0.451 &    
    0.739 &    
    0.641 &    
    0.230 &    
    0.645 &    
    \\
\textbf{\ours-s} &
    57 &    
    0.790 &    
    0.991 &    
    0.418 &    
    0.729 &    
    0.569 &    
    0.203 &    
    0.616 &    
    \\
\textbf{\ours} &
    49 &    
    0.737 &    
    0.975 &    
    0.363 &    
    0.694 &    
    0.550 &    
    0.200 &    
    0.587 &    
    \\
\textbf{\ours} &
    43 &    
    0.672 &    
    0.963 &    
    0.318 &    
    0.649 &    
    0.484 &    
    0.185 &    
    0.545 &    
    \\
\textbf{\ours} &
    39 &    
    0.599 &    
    0.923 &    
    0.270 &    
    0.601 &    
    0.413 &    
    0.168 &    
    0.495 &    
    \\
\midrule
\textcolor{gray}{PixArt-$\Sigma$\cite{pixart_sigma}} 
    & \color{gray} 100  
    & \color{gray} 0.616  
    & \color{gray} 0.984  
    & \color{gray} 0.193  
    & \color{gray} 0.801  
    & \color{gray} 0.478  
    & \color{gray} 0.138  
    & \color{gray} 0.535  
    \\
\midrule
\textbf{\ours} &
    90 &    
    0.629 &    
    0.984 &    
    0.235 &    
    0.806 &    
    0.484 &    
    0.120 &    
    0.543 &    
    \\
\textbf{\ours} &
    80 &    
    0.616 &    
    0.975 &    
    0.198 &    
    0.803 &    
    0.453 &    
    0.123 &    
    0.528 &    
    \\
\textbf{\ours} &
    70 &    
    0.619 &    
    0.984 &    
    0.235 &    
    0.814 &    
    0.438 &    
    0.103 &    
    0.532 &    
    \\
\textbf{\ours} &
    60 &    
    0.606 &    
    0.988 &    
    0.205 &    
    0.806 &    
    0.460 &    
    0.105 &    
    0.528 &    
    \\
\textbf{\ours} &
    50 &    
    0.538 &    
    0.941 &    
    0.190 &    
    0.769 &    
    0.447 &    
    0.093 &    
    0.496 &    
    \\
\bottomrule
\end{tabularx}
}
\smallskip
\caption{\textbf{Detailed quantitative results on the object focused benchmark GenEval \cite{GenEval}.} For FLUX.dev \cite{flux2024}, compression down to moderate parameter budgets (57\,\%) mainly degrades performance on object counting, while other tasks remain relatively stable. In contrast, PixArt-based compression primarily affects positional generation, while performance on other tasks remains largely preserved.}
\label{tab:sup_geneval}
\end{table}
\begin{table}[t]
\centering
{%
\tinyscript
\begin{tabularx}{.75\textwidth}{@{}XS[table-format=3]S[table-format=1.3]S[table-format=1.3]S[table-format=1.3]S[table-format=1.3]S[table-format=1.3]S[table-format=1.3]@{}}
\toprule
\textbf{Method}&
\multicolumn{1}{c}{Params. [\%]} $\downarrow$&
\multicolumn{1}{c}{Concept Art}  $\uparrow$&
\multicolumn{1}{c}{Photo}  $\uparrow$&
\multicolumn{1}{c}{Anime}  $\uparrow$&
\multicolumn{1}{c}{Painting}  $\uparrow$&
\multicolumn{1}{c}{Overall}  $\uparrow$\\
\midrule
\textcolor{gray}{FLUX.dev\cite{flux2024}} 
    & \color{gray} 100  
    & \color{gray} 31.56  
    & \color{gray} 30.26  
    & \color{gray} 33.04  
    & \color{gray} 31.93  
    & \color{gray} 31.70  
    \\
\midrule
\textbf{\ours} &
    90 &    
    31.01 &    
    30.09 &    
    32.73 &    
    31.56 &    
    31.35 &    
    \\
\textbf{\ours} &
    80 &    
    31.00 &    
    30.18 &    
    32.74 &    
    31.64 &    
    31.39 &    
    \\
\textbf{\ours-m} &
    68 &    
    30.94 &    
    29.97 &    
    32.62 &    
    31.53 &    
    31.27 &    
    \\
\textbf{\ours-s} &
    57 &    
    31.00 &    
    29.90 &    
    32.53 &    
    31.45 &    
    31.22 &    
    \\
\textbf{\ours} &
    49 &    
    30.82 &    
    29.86 &    
    32.21 &    
    31.14 &    
    31.01 &    
    \\
\textbf{\ours} &
    43 &    
    29.70 &    
    29.25 &    
    30.99 &    
    30.15 &    
    30.02 &    
    \\
\textbf{\ours} &
    39 &    
    29.28 &    
    28.73 &    
    30.35 &    
    29.61 &    
    29.49 &    
    \\
\midrule
\textcolor{gray}{PixArt-$\Sigma$\cite{pixart_sigma}} 
    & \color{gray} 100  
    & \color{gray} 30.45  
    & \color{gray} 29.13  
    & \color{gray} 31.90  
    & \color{gray} 30.50  
    & \color{gray} 30.50  
    \\
\midrule
\textbf{\ours} &
    90 &    
    29.77 &    
    28.82 &    
    31.37 &    
    29.88 &    
    29.96 &    
    \\
\textbf{\ours} &
    80 &    
    29.57 &    
    28.72 &    
    31.18 &    
    29.67 &    
    29.79 &    
    \\
\textbf{\ours} &
    70 &    
    29.41 &    
    28.61 &    
    31.04 &    
    29.50 &    
    29.64 &    
    \\
\textbf{\ours} &
    60 &    
    29.22 &    
    28.25 &    
    30.74 &    
    29.17 &    
    29.34 &    
    \\
\textbf{\ours} &
    50 &    
    28.05 &    
    27.42 &    
    29.68 &    
    27.96 &    
    28.28 &    
    \\
\bottomrule
\end{tabularx}
}
\smallskip
\caption{\textbf{Detailed quantitative results on HPSv2\cite{HPSv2}.} \ours shows consistent and graceful degradation across all categories, without any single category degrading disproportionately.}
\label{tab:sup_hps}
\end{table}
\begin{table}[t]
\centering
{%
\tinyscript
\begin{tabularx}{.85\textwidth}{@{}XS[table-format=3]S[table-format=2.1]S[table-format=2.1]S[table-format=2.1]S[table-format=2.1]S[table-format=2.1]S[table-format=2.1]S[table-format=2.1]S[table-format=2.1]@{}}
\toprule
\textbf{Method}&
\multicolumn{1}{c}{Params. [\%]}  $\downarrow$&
\multicolumn{1}{c}{Global}  $\uparrow$&
\multicolumn{1}{c}{Entity}  $\uparrow$&
\multicolumn{1}{c}{Attribute}  $\uparrow$&
\multicolumn{1}{c}{Relation}  $\uparrow$&
\multicolumn{1}{c}{Other}  $\uparrow$&
\multicolumn{1}{c}{Overall}  $\uparrow$\\
\midrule
\textcolor{gray}{FLUX.dev\cite{flux2024}} 
    & \color{gray} 100  
    & \color{gray} 83.3  
    & \color{gray} 90.1  
    & \color{gray} 86.8  
    & \color{gray} 93.5  
    & \color{gray} 82.8  
    & \color{gray} 83.9  
    \\
\midrule
PPCL\cite{PPCL} &
    68 &    
    85.1 &    
    87.9 &    
    85.6 &    
    89.9 &    
    87.8 &    
    80.0 &    
    \\
\textbf{\ours} &
    90 &    
    83.6 &    
    89.9 &    
    87.0 &    
    93.4 &    
    83.2 &    
    83.6 &    
    \\
\textbf{\ours} &
    80 &    
    82.4 &    
    89.9 &    
    86.8 &    
    93.2 &    
    82.8 &    
    83.6 &    
    \\
\textbf{\ours-m} &
    68 &    
    81.8 &    
    89.2 &    
    86.7 &    
    93.0 &    
    79.6 &    
    83.4 &    
    \\
\textbf{\ours-s} &
    57 &    
    82.1 &    
    89.1 &    
    86.5 &    
    92.5 &    
    80.0 &    
    82.6 &    
    \\
\textbf{\ours} &
    49 &    
    82.1 &    
    89.1 &    
    86.4 &    
    92.3 &    
    79.6 &    
    81.7 &    
    \\
\textbf{\ours} &
    43 &    
    79.0 &    
    87.6 &    
    85.3 &    
    92.1 &    
    79.2 &    
    79.9 &    
    \\
\textbf{\ours} &
    39 &    
    79.6 &    
    86.0 &    
    84.6 &    
    89.9 &    
    77.2 &    
    78.1 &    
    \\
\midrule
\textcolor{gray}{PixArt-$\Sigma$\cite{pixart_sigma}} 
    & \color{gray} 100  
    & \color{gray} 82.1  
    & \color{gray} 85.8  
    & \color{gray} 84.4  
    & \color{gray} 90.1  
    & \color{gray} 72.0  
    & \color{gray} 78.8  
    \\
\midrule
\textbf{\ours} &
    90 &    
    83.9 &    
    85.4 &    
    84.8 &    
    90.3 &    
    74.4 &    
    78.5 &    
    \\
\textbf{\ours} &
    80 &    
    83.6 &    
    85.2 &    
    84.4 &    
    90.5 &    
    74.0 &    
    78.0 &    
    \\
\textbf{\ours} &
    70 &    
    84.5 &    
    85.1 &    
    83.9 &    
    90.8 &    
    74.0 &    
    78.2 &    
    \\
\textbf{\ours} &
    60 &    
    84.2 &    
    84.9 &    
    84.8 &    
    90.7 &    
    74.4 &    
    78.0 &    
    \\
\textbf{\ours} &
    50 &    
    83.6 &    
    84.7 &    
    84.2 &    
    90.17 &    
    72.0 &    
    76.6 &    
    \\
\bottomrule
\end{tabularx}
}
\smallskip
\caption{\textbf{Detailed quantitative results on DPG\cite{DPG}.} \ours maintains competitive performance across all DPG categories with graceful degradation under stronger compression.}
\label{tab:sup_dpg}
\end{table}
Our \ours successfully compresses both large and small transformer models with minimal quality degradation, achieving reductions down to 50\,\% of the original parameter budget with only around a 5\,\% average performance drop across the GenEval~\cite{GenEval}, DPG~\cite{DPG}, and HPSv2~\cite{HPSv2} benchmarks. Detailed results in \cref{tab:sup_geneval,tab:sup_dpg,tab:sup_hps} further show balanced degradation across all subcategories, with no task or category emerging as a systematic failure mode.

\textbf{Compatibility with quantisation.}
Quantisation is complementary to \ours and can be applied on top of the compressed models. Using simple per-tensor FP8 quantisation, the average quality reduction $R$ increases from 0.75 to only 1.87 for \ours-m and from 2.60 to 3.57 for \ours-s, indicating that the compressed models remain robust under additional quantisation.

\section{Training Specifications}
\label{sup:training_specs}

\begin{table}[t]
\centering
{%
\tinyscript
\begin{tabularx}{.6\textwidth}{@{}
S[table-format=3.0, table-column-width=1.15cm] 
    S[table-format=2.0, table-column-width=1.15cm] 
    S[table-format=5.0, table-column-width=1.15cm] 
    S[table-format=2.0, table-column-width=1.15cm] 
    S[table-format=1.2, table-column-width=1.15cm] 
    S[table-format=2.2, table-column-width=1.15cm] 
@{}}
\toprule
\textbf{start $P$}&
\textbf{end $P$}&
\textbf{Steps}&
\textbf{$K$}&
\textbf{$c_m$}&
\textbf{$\alpha$}
\\
\midrule
\multicolumn{6}{c}{FLUX.dev\cite{flux2024}}\\
\midrule
    100 &      
    90 &    
    20000 &    
    30 &    
    0.25 &    
    8.30    
    \\
    90 &      
    80 &    
    20000 &    
    30 &    
    0.25 &    
    5.94     
    \\
    100 &      
    68 &    
    40000 &    
    48 &    
    0.60 &    
    12.70     
    \\
    68 &      
    57 &    
    20000 &    
    30 &    
    0.35 &    
    4.58     
    \\
    57 &      
    49 &    
    20000 &    
    30 &    
    0.40 &    
    9.38     
    \\
    49 &      
    43 &    
    20000 &    
    30 &    
    0.40 &    
    12.20     
    \\
    43 &      
    39 &    
    20000 &    
    30 &    
    0.30 &    
    8.79     
    \\
\midrule
\multicolumn{6}{c}{PixArt-$\Sigma$\cite{pixart_sigma}}\\
\midrule
    100 &      
    90 &    
    88500 &    
    21 &    
    0.20 &    
    4.30     
    \\
    90 &      
    80 &    
    88500 &    
    21 &    
    0.20 &    
    1.60     
    \\
    80 &      
    70 &    
    88500 &    
    21 &    
    0.40 &    
    18.00     
    \\
    70 &      
    60 &    
    88500 &    
    21 &    
    0.40 &    
    13.00     
    \\
    60 &      
    50 &    
    88500 &    
    21 &    
    0.40 &    
    8.97    
    \\
\bottomrule
\end{tabularx}
}
\smallskip
\caption{\textbf{Training details.} For every training of \ours from a start parameter budget to an end parameter budget $P_{start}\longrightarrow P_{end}$ we provide the number of compressed blocks $K$, with the maximum compression $c_m$, the compression slope $\alpha$, and the number of training steps. For readability, we provide $\alpha$ scaled by 1000.}
\label{tab:sup_training}
\end{table}
In \cref{tab:sup_training}, we report the training hyperparameters of \ours for different compression levels, including the maximum compression $c_m$ and the slope $\alpha$. Higher compression levels are obtained through an iterative procedure in which models trained at lower compression levels are further compressed and retrained.

For SVDtrunc, the probing stage requires approximately  175 (parallelisable) H200 GPU hours, while distillation adds  126 GPU hours for SVDtrunc-m and additional 63 GPU hours for the SVDtrunc-s model.

\section{Ablations}
\label{sup:ablations}
\begin{figure}
    \label{tab:abl_uniform_vs_ours}
    \centering
    \includegraphics[width=.7\linewidth]{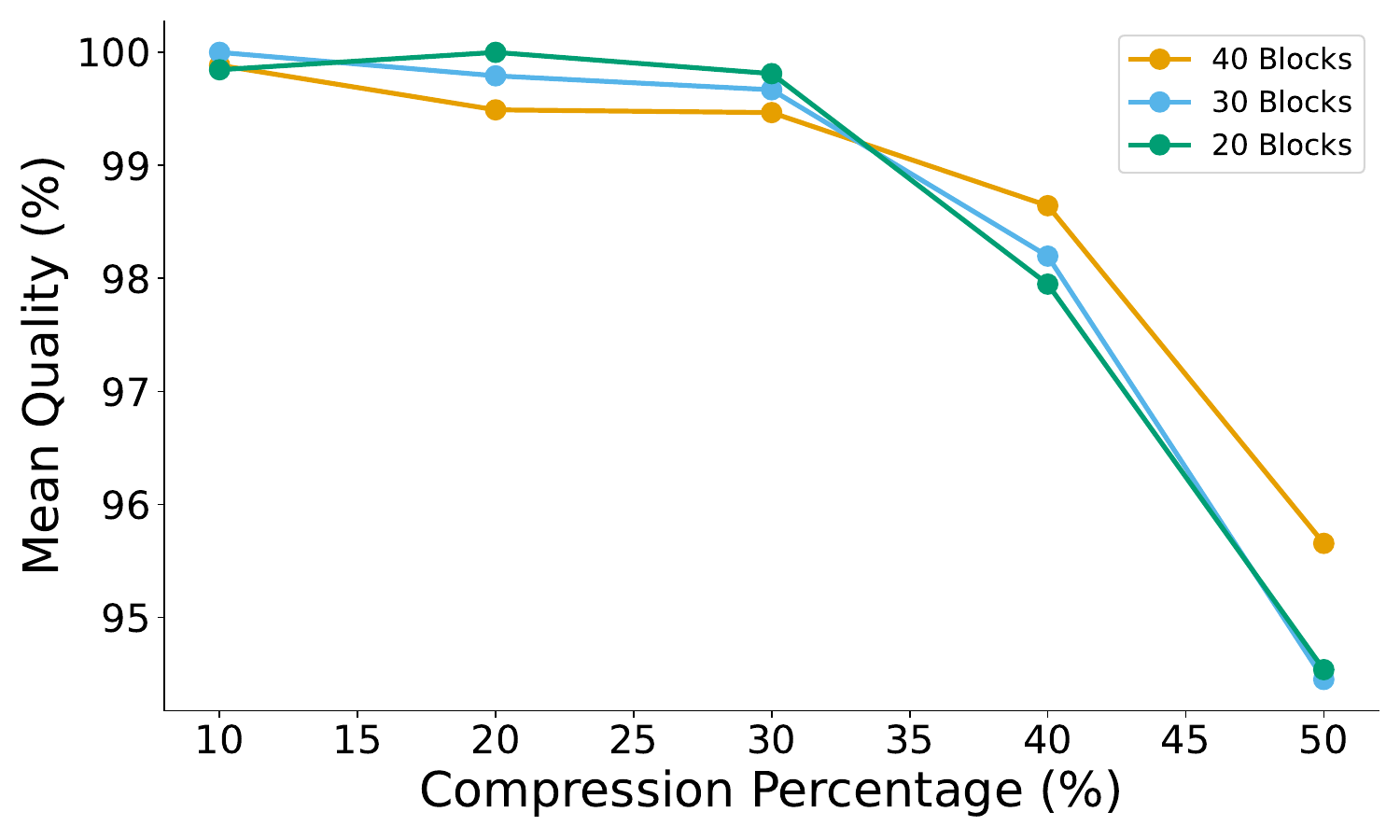}
    \captionof{figure}{\textbf{Quality–parameter trade-off} for different numbers of compressed blocks.}
    \label{fig:ablation_depth}
\end{figure}

We conduct two additional ablation studies to analyse the factors influencing performance of \ours. We examine the effect of the distribution of compression across blocks and the choice of fine-tuning objective.

\textbf{Influence of the Number of Compressed Blocks.}
We analyse the influence of the number of compressed blocks by varying $|\mathbb{S}| \in \{20, 30, 40\}$ under comparable global parameter budgets (\cref{fig:ablation_depth}). 
At moderate compression levels (20–30\,\%), performance differences are negligible (below one percentage point) across all configurations. As compression becomes more aggressive (40–50\,\%), distributing rank reduction across more blocks yields improved quality retention compared to compressing fewer blocks more severely. 
These results indicate that redundancy is distributed throughout the Transformer backbone. This means that mild compression is largely insensitive to depth or the number of compressed blocks, whereas under stronger constraints, broader distribution better preserves performance.

\textbf{Influence of the Fine-Tuning Objective.}
We compare fine-tuning with \emph{(1)} the rectified-flow objective only, \emph{(2)} the modular knowledge distillation (MKD) only, and \emph{(3)} the combined objective at a 68\,\% parameter budget.
Fine-tuning with a rectified-flow objective alone results in a higher overall performance decrease of $R=2.12\,\%$. The MKD-only and the combination of both perform roughly on a par with an averaged quality reduction of $R=0.64\,\%$ and $R=0.93\,\%$, respectively.
We use the combined objective in all main experiments to preserve consistency with the native rectified-flow training objective, while maintaining competitive benchmark performance under compression.

\section{Domain Drift and Memorisation loss}
\label{sup:domain_drift}

As described in the main paper, we occasionally observe a domain drift in the generated outputs under strong compression. Interestingly, this drift does not necessarily correspond to a reduction in visual quality. Instead, the generated images move towards slightly different stylistic domains while remaining visually coherent.

\begin{figure}
    \centering
    \includegraphics[width=.9\linewidth]{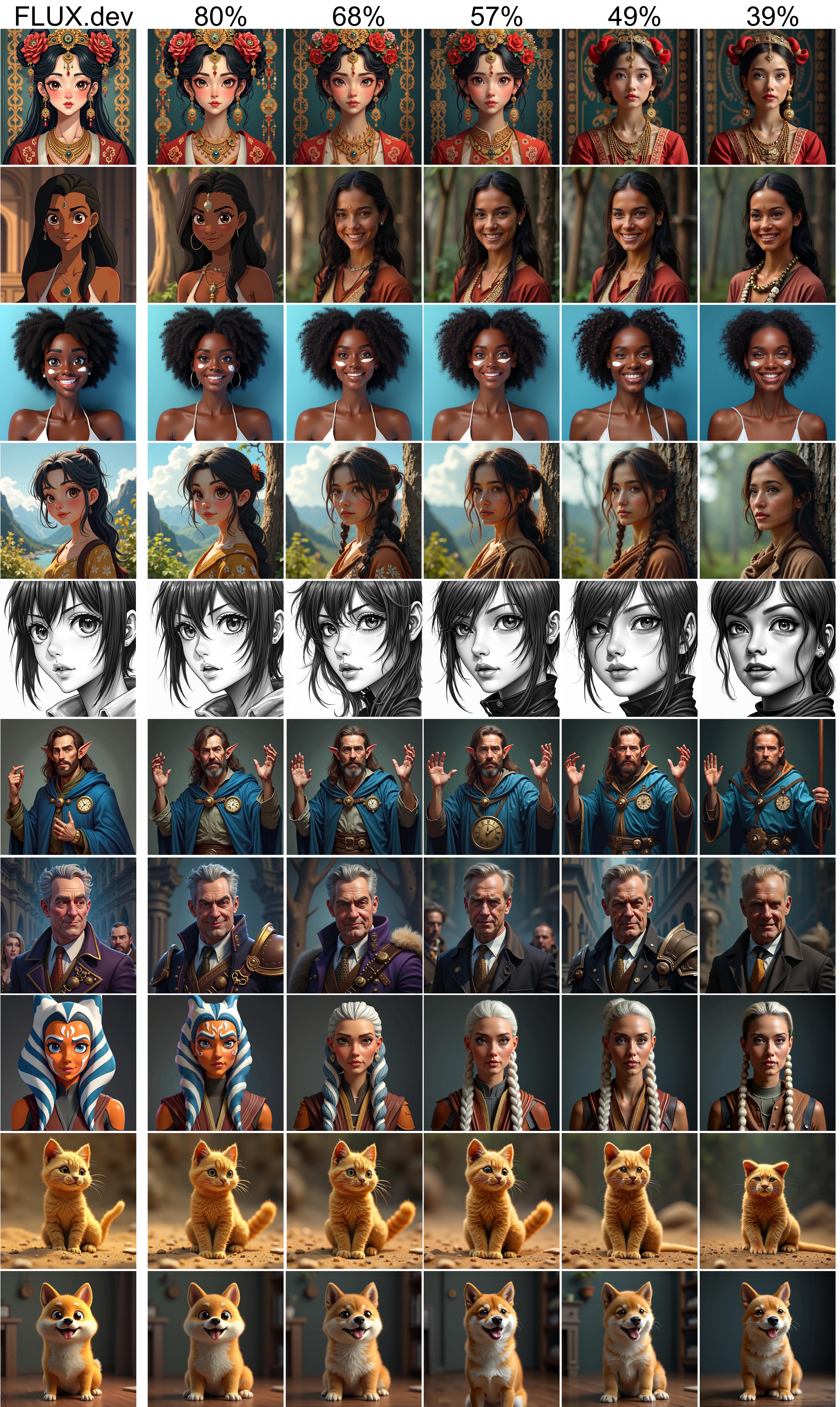}
    \caption{\textbf{Domain drift examples.} With aggressive compression, the model can diverge from manga or cartoonish styles towards more realistic images.}
    \label{fig:sup_domain_drift}
\end{figure}

In \cref{fig:sup_domain_drift}, we present additional qualitative examples for models compressed between 80\,\% and 39\,\% parameter budgets of the FLUX.dev~\cite{flux2024} model. The drift appears gradually as compression increases and often manifests as an alignment of stylised domains, such as anime or cartoon-like images, towards more realistic appearances.

\begin{figure}
    \centering
    \includegraphics[width=.9\linewidth]{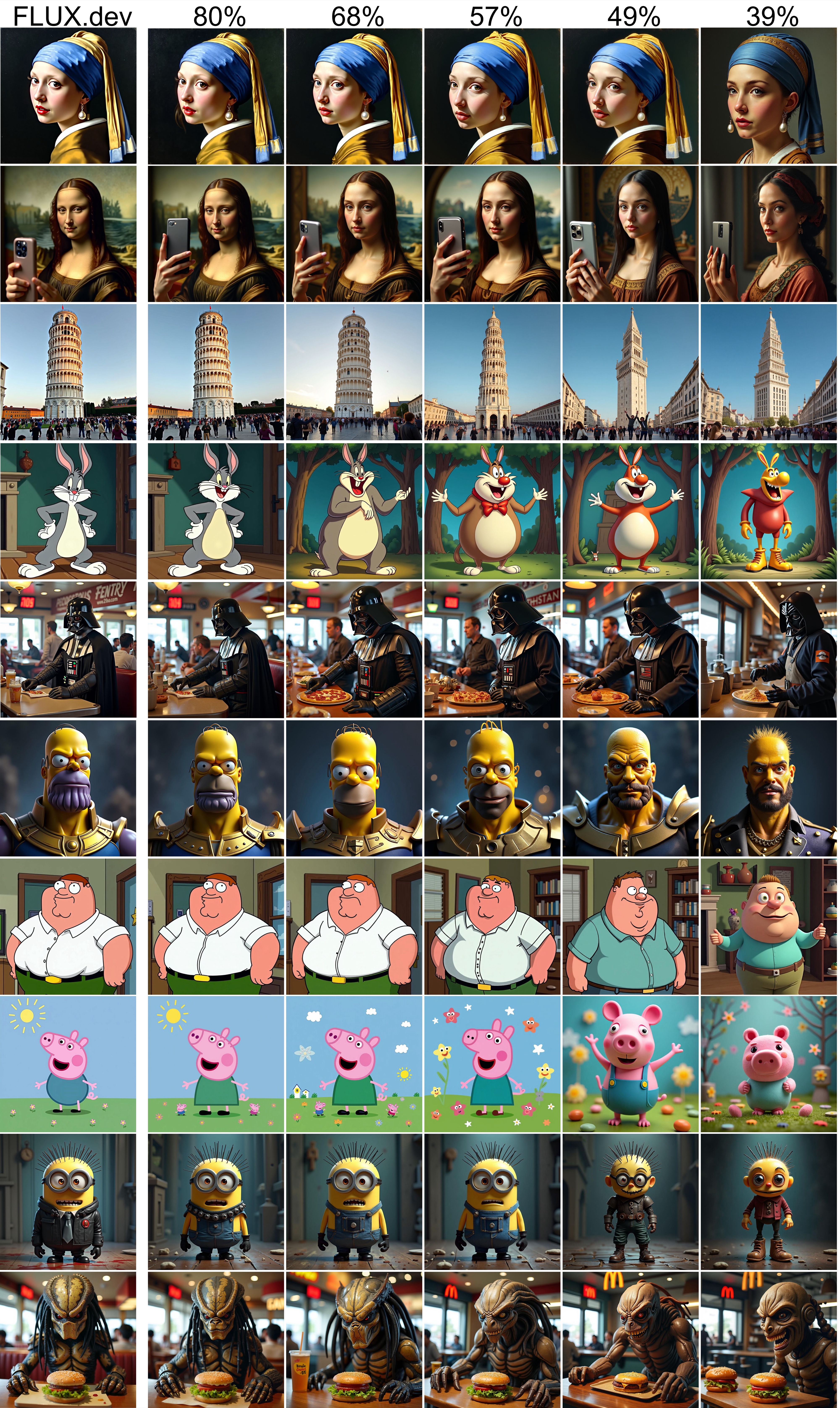}
    \caption{\textbf{Memorisation loss examples.} With aggressive compression, the model can not reproduce iconic instances like the tower of Pisa or Bugs Bunny.}
    \label{fig:sup_memory_shift}
\end{figure}

In \cref{fig:sup_memory_shift}, we present additional qualitative results for the loss of memorised specific concepts under increasing compression rates. Under strong compression, parameters that support instance-specific memorisation may be less preserved than those encoding more broadly useful image-synthesis capabilities.

\section{Additional Qualitative Examples}
\label{sup:add_qual_examples}

\begin{figure}
    \centering
    \includegraphics[width=.9\linewidth]{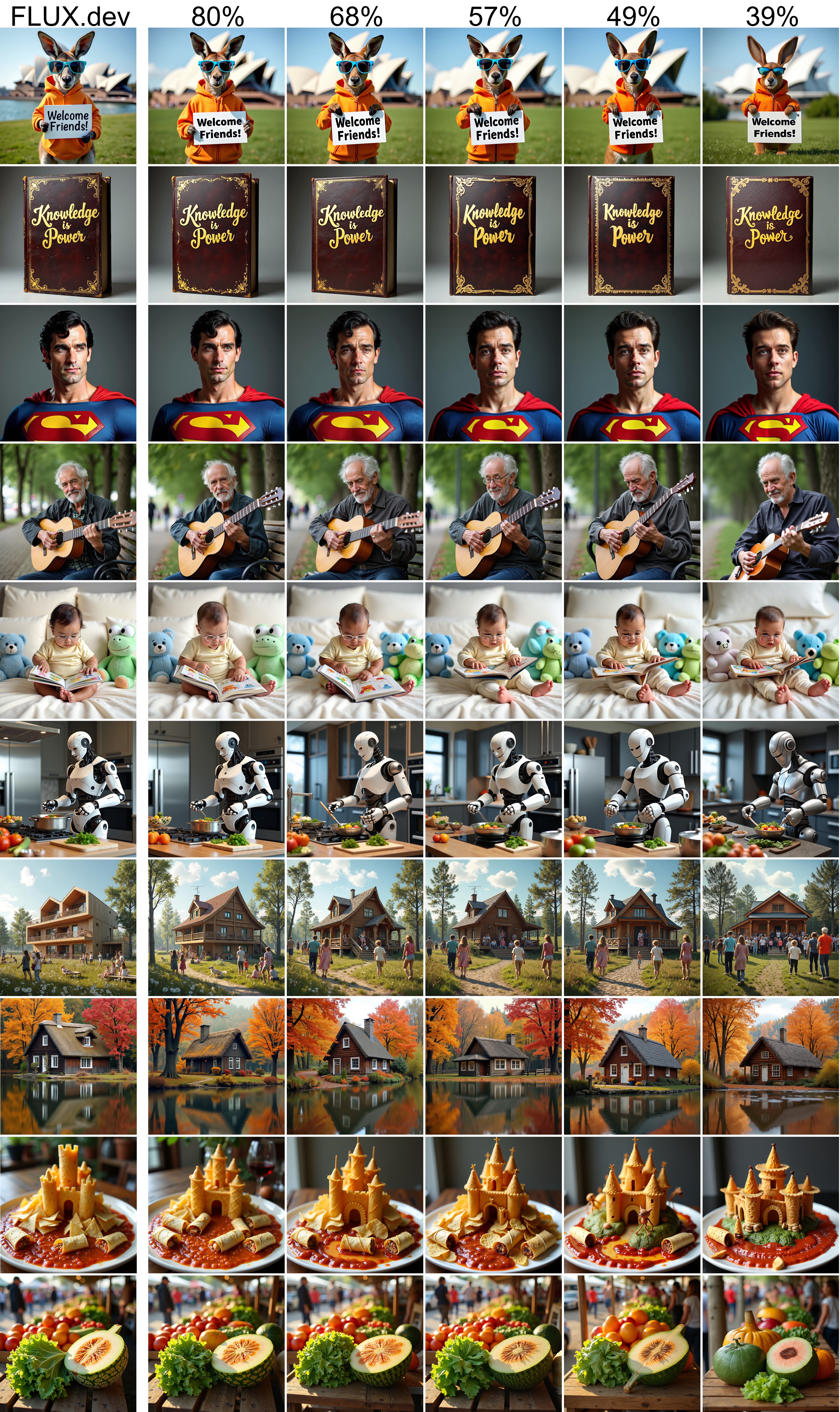}
    \caption{\textbf{Additional qualitative examples.}}
    \label{fig:sup_qual_examples_1}
\end{figure}

\begin{figure}
    \centering
    \includegraphics[width=.9\linewidth]{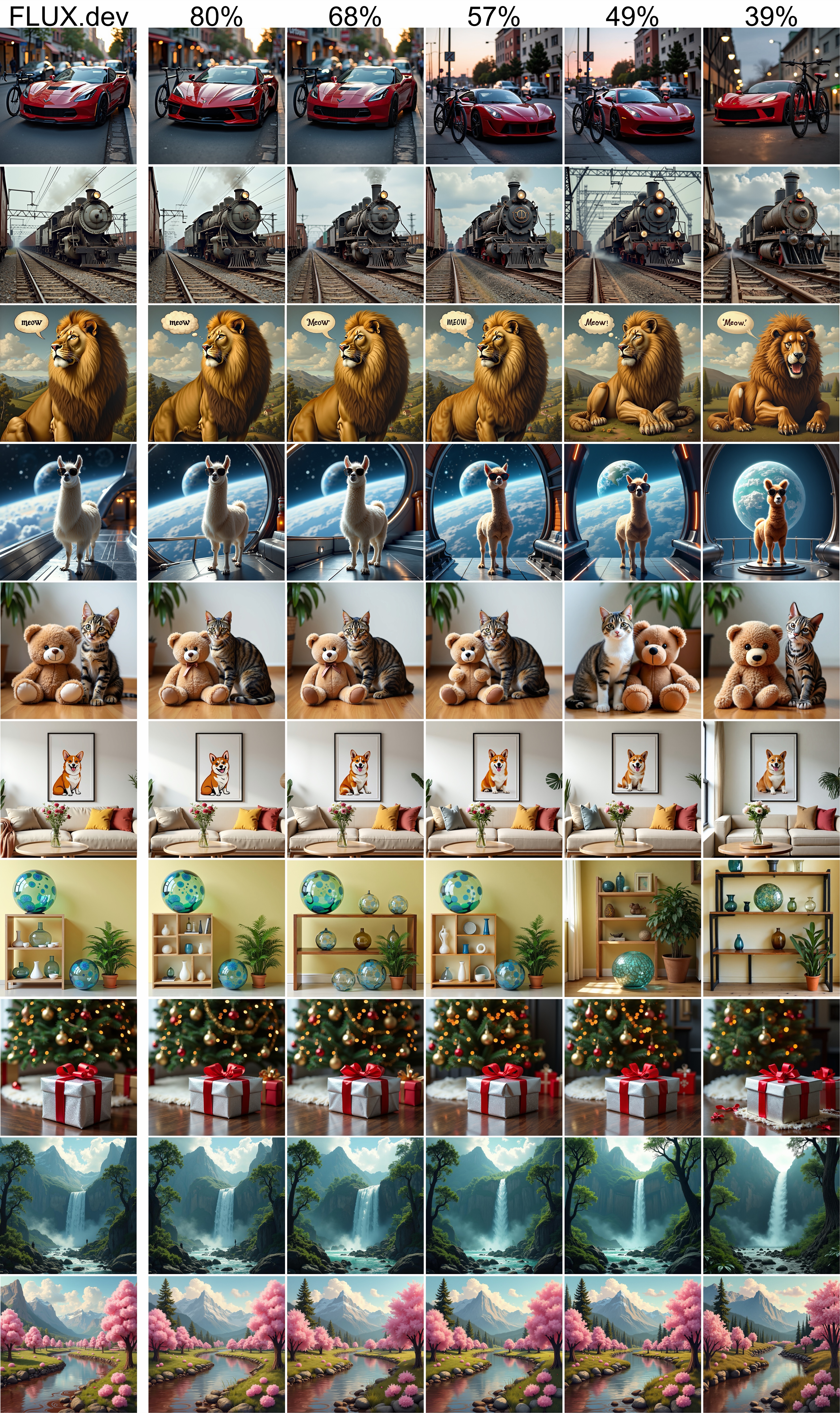}
    \caption{\textbf{Additional qualitative examples.}}
    \label{fig:sup_qual_examples_2}
\end{figure}

We present additional qualitative examples across different parameter budgets ranging from 80\,\% to 39\,\% in \cref{fig:sup_qual_examples_1,fig:sup_qual_examples_2}. Even under strong compression, the decrease in generative quality remains gradual and visually subtle. The compressed models continue to produce high-quality images with diverse compositions, exhibiting only minor losses in fine details at more aggressive compression levels. This observation is consistent with the quantitative results, which show a gradual and modest decrease in overall generative performance as compression increases (\cref{sup:add_quant_results}).

\end{document}